\documentclass{article}

\usepackage{arxiv}
\usepackage[utf8]{inputenc}
\usepackage[T1]{fontenc}
\usepackage[hidelinks]{hyperref}       
\usepackage{url}            
\usepackage{booktabs}       
\usepackage{amsfonts}       
\usepackage{nicefrac}       
\usepackage{microtype}      
\usepackage{lipsum}
\usepackage{graphicx}
\usepackage{makecell}
\usepackage{subcaption}
\usepackage{amsmath}
\usepackage{ragged2e}
\usepackage{caption}
\usepackage{multirow}
\usepackage{adjustbox}
\usepackage{accents}
\usepackage[numbers]{natbib}
\usepackage{tcolorbox}
\usepackage{listings}
\usepackage{multibib}
\newcites{primary}{Primary Sources}

\title{Multimodal LLMs for OCR, OCR Post-Correction, and Named Entity Recognition in Historical Documents}

\author{%
\textbf{Gavin Greif}\thanks{These authors share first-author contribution.}\\
Centre for Economic and Social History\\
University of Oxford\\
\texttt{gavin.greif@history.ox.ac.uk}
\and
\textbf{Niclas Griesshaber}\footnotemark[1]\\
Department of Economics\\
University of Mannheim\\
\texttt{niclas.griesshaber@uni-mannheim.de}
\\[0.8cm] 
\and
\textbf{Robin Greif}\\
Department of Theoretical Physics\\
University of Oxford\\
\texttt{robin.greif@physics.ox.ac.uk}
}

\newcommand{\olde}[1]{\accentset{\scriptscriptstyle e}{#1}}

\begin{document}

\maketitle

\begin{abstract}
We explore how multimodal Large Language Models (mLLMs) can help researchers transcribe historical documents, extract relevant historical information, and construct datasets from historical sources. Specifically, we investigate the capabilities of mLLMs in performing (1) Optical Character Recognition (OCR), (2) OCR Post-Correction, and (3) Named Entity Recognition (NER) tasks on a set of city directories published in German between 1754 and 1870. First, we benchmark the off-the-shelf transcription accuracy of both mLLMs and conventional OCR models. We find that the best-performing mLLM model significantly outperforms conventional state-of-the-art OCR models and other frontier mLLMs. Second, we are the first to introduce multimodal post-correction of OCR output using mLLMs. We find that this novel approach leads to a drastic improvement in transcription accuracy and consistently produces highly accurate transcriptions ($<$1\% CER), without any image pre-processing or model fine-tuning. Third, we demonstrate that mLLMs can efficiently recognize entities in transcriptions of historical documents and parse them into structured dataset formats.  Our findings provide early evidence for the long-term potential of mLLMs to introduce a paradigm shift in the approaches to historical data collection and document transcription. 
\end{abstract}

\section{Introduction}

The rate at which researchers can collect historical evidence is constrained by their ability to access the relevant sources, accurately transcribe them, and extract the desired information. While a global effort to scan historical sources is gradually reducing access barriers, the need for manual labor in the transcription and extraction of historical information remains a major bottleneck. To remedy this, researchers have come up with a wide range of (semi-)automated solutions to transcribe images of historical documents and extract the desired information from the generated transcriptions. However, existing solutions are either inaccurate or require significant technical expertise and manual annotations. In this paper, we introduce a new approach, which addresses these existing problems by leveraging multimodal Large Language Models (mLLMs). Our approach accurately transcribes historical documents and efficiently parses the information they contain into structured datasets. 

Since it is impossible to exhaustively benchmark our approach for all languages, fonts, and layouts found in historical sources in a single paper, we focus our attention on historical Latin-alphabet prints. While this may seem restrictive, by 1800, already over 1.7 million books that fall into this category had been published across Europe \cite{buringh2009charting}. The rates of book production further increased across the nineteenth and twentieth centuries, and by 1910, Germany alone was publishing well over 30,000 new titles every year, almost as many as France, England, and the US combined \cite{tatlock2010introduction}. Importantly, Latin-alphabet Europe was divided until the mid-twentieth century, with German-speaking and Lutheran Europe predominantly using Blackletter fonts (e.g., Fraktur) and the rest using Roman fonts (e.g., Antiqua) \cite{kaprFrakturFormUnd1993}. While automating the accurate transcription is undoubtedly easier for printed than handwritten texts, historical typefaces, especially Blackletter ones, have posed a significant challenge to automating historical document transcription \cite{reulStateArtOptical2018a}. 

To demonstrate how exactly our generalized approach outperforms existing solutions, we use a corpus of thirty pages drawn from city directories published across German-speaking Europe between 1754--1870. City, trade, and business directories are an important source for economists, historians, and genealogists, as they offer individual-level micro-data for periods and places where alternative sources (such as censuses) are not available \cite{greifMerchantsProtoFirmsGerman2022}. For German-speaking Europe alone, well over 40,000 surviving directories, each containing thousands of entries, have been located \cite{Junkers2023}. Yet, this cornucopia of historical data has, thus far, not been economic to transcribe. The thirty pages in our corpus include the typical challenges associated with the machine transcription of historical printed texts, including multiple historical typefaces, excess visual information in the scans, skewed texts, handwritten marginalia, discolored backgrounds, print errors, different historical fonts on the same page, and various visual artifacts. All pages in the corpus were manually transcribed and corrected several times to generate the ground truth for the evaluation of our approach. 

We make three contributions in this paper. First, we benchmark the transcription capabilities of mLLMs. We compare the transcription performance of two leading mLLMs (Gemini 2.0 Flash; GPT-4o) against the leading off-the-shelf models for nineteenth century historical prints (Transkribus' Text Titan I and Print M1) and a widely used conventional OCR engine (Tesseract 5.5.0, "deu\_frak"). We find that in a one-shot setting, Gemini 2.0 Flash achieves the highest accuracy out of all tested models and that its off-the-shelf transcription accuracy is comparable to that of corpus-fine-tuned OCR models reported in the literature for similar documents. Second, we are the first to introduce mLLM OCR Post-Correction for historical texts. We show that by combining the original source image and a noisy transcription, mLLMs can substantially increase the accuracy of OCR output, thus demonstrating that, by focusing on text-only post-correction, existing research has missed the full potential of mLLMs. Finally, we explore the capabilities of mLLMs in recognizing entities in our irregularly structured historical sources. We provide early evidence indicating that mLLMs offer an accessible and efficient solution to identifying and parsing historical information. We find that, while a direct parsing of information from images into the dataset is possible, separating out the steps and including mLLM-based OCR Post-Correction yields the highest accuracy. We release all code and data used in this paper.\footnote{Available at: \url{https://github.com/niclasgriesshaber/llm_historical_dataset_benchmarking.git} and \\ \url{https://github.com/niclasgriesshaber/gemini_historical_dataset_pipeline.git}}

\section{Related Work}

\subsection{The Traditional OCR Pipeline}

While research on automating the transcription of historical documents is vast, existing research typically focuses on one or multiple parts of what we call the \textit{traditional pipeline}. This standard workflow follows a sequential process, consisting of separate pre-processing, layout recognition, character recognition, and post-processing stages. Depending on the project, a separate Named Entity Recognition (NER) stage may be employed to classify information contained in the post-processed transcription. While existing work on this \textit{traditional pipeline} is too vast to be discussed here in full, a brief review of it is necessary to highlight the challenges of automating historical document transcriptions, the existing approaches to resolving them, and the way in which mLLMs have been used in this context.

The first step of the traditional pipeline is the pre-processing of the images of the historical source. If present, excess visual information, such as parts of another page, the book frame, dark spaces, or an archive footer, are removed during this step. Moreover, the image may be de-skewed and processed through denoising and gray-scaling or binarization to remove unwanted artifacts and increase the contrast between the text and the page background \cite{guptaOCRBinarizationImage2007,holley2009good,sulaiman2019degraded}. For cases in which images are degraded, up-sampling has also been applied during pre-processing \cite{TasPreProcessing}. Despite many OCR engines having built-in pre-processing which automatically carries out some of these image manipulations and some researchers using tools such as You Only Look Once (YOLO) to automate some of the pre-processing \cite{Constum2024EndToEnd}, recent scholarship continues to rely on extensive tool-assisted manual pre-processing of source images \cite{albersPerksPitfallsCity2023}. 

The second step in the traditional pipeline is the recognition of the document layout. This involves, inter alia, image segmentation, line recognition, baseline detection, and reading order determination. While most modern transcription engines have built-in layout recognition capabilities that work well for single-column texts, more complex layouts often require a separate layout recognition step to ensure the correct reading order. The tools and approaches to automated layout recognition are many. However, what matters for the present paper is that, at the time of writing, the accurate recognition of complex layouts typically requires the custom fine-tuning of a model using significant amounts of annotated data or manual post-correction \cite{reulOCR4allAnOpenSourceTool2019,fleischhackerImprovingOCRQuality2024,shenLayoutParserUnifiedToolkit2021,gutehrleProcessingStructureDocuments2022,dobranicEstimatingNumberAnnotations2024,clericeYouActuallyLook2023}.

The third step in the traditional pipeline is the recognition of the characters in the image. For historical sources, this is particularly challenging as there are an abundance of different historical typefaces, languages, and fonts. Conventional OCR engines such as Kraken, PyLaia, PeroOCR, or Calamari-OCR use a Convolutional Neural Networks (CNNs) for feature extraction, Bidirectional Long Short-Term Memory (BiLSTM) networks for sequence modeling, and Connectionist Temporal Classification (CTC) for decoding. Recently, this CNN-BiLSTM infrastructure has received competition from solutions that employ a transformer-based architecture such as TrOCR \cite{Stroebel2023}. These allow self-supervised pre-training and have especially been used for recognizing historical handwriting \cite{ströbel2022transformerbasedhtrhistoricaldocuments}. However, to achieve the highest transcription accuracies, existing solutions generally require corpus-specific fine-tuning \cite{springmann2017ocr, strobel2023adaptability}. 

Finally, since character recognition solutions usually provide imperfect transcriptions, researchers generally apply one or more post-processing steps to further improve transcription accuracy. Traditional OCR Post-Correction approaches include, inter alia, manual post-correction, rules-based correction (e.g., dictionary-based), and probabilistic models (e.g., Hidden Markov Models) \cite{xu2017retrieving,cao2017us,nguyenSurveyPostOCRProcessing2021}. However, as in other parts of the traditional pipeline, deep-learning methods have become state-of-the-art in the post-correction of transcriptions of historical documents. While sequence-to-sequence models have been used for post-correction \cite{duong2021unsupervised}, most approaches currently use transformer-based architectures, typically based on BERT, T5, or their derivatives \cite{Chen2023,Dereza2024, hajiali2023ocr, Wu2023, Dereza2024b,Wolters2024, nguyenNeuralMachineTranslation2020,lofgren-dannells-2024-post}. Recent work has sought to further improve OCR Post-Correction by incorporating OCR and its post-correction to increase accuracy by using more contextual information \cite{bourne2025clocrccontextleveragingocr, ChenStrobel2024}.

Researchers often want to classify or extract relevant information in their transcriptions. For this, BERT-based NER approaches are currently the standard solution. These have been found to be better at handling OCR noise, archaic spelling variations, and complex entity structures, and have been shown to outperform both Conditional Random Field (CRF) and BiLSTM-CRF based approaches \citet{ehrmann2024named}. In NER too, the best results are typically achieved by fine-tuning a model to the specific dataset \cite{tualBenchmarkNestedNamed2023}. 

To enhance efficiency and accessibility, researchers often integrate multiple steps of this pipeline or provide a full end-to-end solution. These solutions can be both modular and integrated. For example, \citet{petitpierreEndtoendPipelineHistorical2023} introduce a five-phase modular end-to-end pipeline consisting of layout analysis, text segment extraction, Handwritten Text Recognition, tabular structuring, and post-processing. Similarly, \citet{vankoertLoghiEndtoEndFramework2024} introduce a three-stage modular pipeline consisting of a layout analysis stage that uses Laypa for layout segmentation, an HTR stage built on Keras, and a post-correction stage. Simultaneously, various integrated end-to-end pipelines have been proposed with some, but not all, including NER capabilities \cite{Rouhou2022Transformer, constum2025daniel, coquenet2023dan, hamdan2024hand, Castro2024dancer}. Transkribus and OCR4All offer two user-friendly and accessible options for non-technical users seeking an end-to-end solution with comparatively easy ways to manually post-correct at various steps of the process \cite{reulOCR4allAnOpenSourceTool2019,kahleTranskribusaServicePlatform2017,neudecker2019ocr}.  

\subsection{Multimodal Large Language Models in Historical Document Transcription}

Despite the rapid development of mLLMs, the existing research on their applications for historical document transcription is limited. To our knowledge, there currently exists no work on their potential in aiding document pre-processing; however, a few projects have sought to test their transcription capabilities for historical documents.  \citet{liHandwritingRecognitionHistorical2024} found that for handwritten texts in French, Italian, Spanish, and Dutch published between the sixteenth and nineteenth centuries, fine-tuned TrOCR and CNN-BiLSTM models drastically outperform an unspecified Gemini model. \citet{kimEarlyEvidenceHow2025} found that for (mostly) handwritten probation records from 1921 Belgium, Claude (prompted with two few-shot examples) produced a more accurate transcription than other OCR engines (EasyOCR, TrOCR, KerasOCR, Tesseract) and outperformed fine-tuned TrOCR versions. \citet{humphriesUnlockingArchivesUsing2024} found that for their corpus of eighteenth and nineteenth century English handwriting, Gemini-1.5-pro, GPT-4o, and Claude-Sonnet-3.5 all achieved transcription accuracies comparable to and sometimes better than conventional state-of-the-art OCR algorithms. \citet{ghiritiExploringCapabilitiesGPT4Vision2024} tested the transcription capabilities of GPT-4 Vision-Preview and its response to various artificially introduced distortions and degradations for a corpus of early twentieth century German-language Fraktur prints and found that it outperformed Tesseract, except for those documents with complex layouts. 

The OCR Post-Correction literature has also seen several contributions discussing the applications of LLMs. However, remarkably, all existing work on mLLMs in OCR Post-Correction is exclusively text based and no existing work has leveraged the visual capabilities of mLLMs for OCR Post-Correction. For modern texts, the post-correction suitability of mLLMs appears undisputed. Working on the modern-font ICDAR 2013 and 2023 datasets, \citet{Chen2023} found that GPT-3.5-turbo-0301 improved the accuracy of noisy OCR output. Similarly, \citet{hajiali2023ocr} found that for a corpus of modern texts, GPT-based post-correction outperformed a BERT-FastText model. For historical texts, however, evidence is more ambiguous. Boros et al. tested various LLaMA, BLOOM(Z), OPT and GPT models on several historical datasets and concluded that ``LLMs are not good at correcting transcriptions of historical documents of any kind, at least in the applied experimental setting. Not only do they not improve the original transcription, they usually degrade them, making LLM-based post-correction of historical transcriptions a rather distant and even dangerous prospect'' \cite[p.141]{Boros2024}. Meanwhile, \citet{Thomas2024} found that, for a corpus of nineteenth-century British newspaper transcriptions, LLaMA 2 outperformed a conventional BART-based post-correction approach. Similarly, \citet{kanervaOCRErrorPostCorrection2025} showed that for English historical material mLLMs achieve notable improvements in post-correcting OCR errors in a zero-shot setting, while for material in historical Finnish, only GPT-4o achieved a marginal improvement. In contrast to this existing scholarship, we include the source image to leverage the information contained within to allow the mLLM to correct formatting and transcription errors. This additional information reduces the task to finding mismatches, thereby reducing the need to guess - or even worse hallucinate - the intended meaning of words in the absence of visual reference information.

When it comes to information extraction and named entity recognition tasks, mLLMs have also received some attention. For historical transcriptions, initial evidence suggested that, early models, such as GPT-3.5, were not yet good at NER tasks \cite{gonzalez-gallardoYesCanChatGPT2023}. \citet{gonzalez-gallardoLeveragingOpenLarge2024} found that while fine-tuned historical NER models (Stacked NER; Temporal NER) clearly outperformed off-the-shelf instruction-tuned models (ChatGPT; LLaMA-chat; Mixtral; Zeyphr), the latter may still be helpful tools for human annotators. Most recently, however, \citet{hiltmannNER4allContextAll2025} have shown that for a nineteenth century German-language source, GPT-4o outperformed off-the-shelf Flair (ner-german-large) and SpaCy (de\_core\_news\_lg) in NER tasks. While these results appear promising, the authors did not compare mLLMs to a corpus-fine-tuned model. \citet{xieMultimodalLLMassistedInformation2025} use GPT-4o to extract and classify relevant information in Swedish Patent Cards (1945-1975) directly from the image, yet offer no comparison in accuracy to alternative approaches. At the time of writing, BERT-based approaches continue to be dominant in recognizing entities in historical transcriptions and NER tasks are frequently limited to simple recognition of places and locations in a text \cite{gruberMultilabelClassificationNamed2025}. However, in other domains such as medicine, mLLMs have proven to outperform or be comparable to human annotators in extracting and parsing relevant information \cite{balasubramanianLeveragingLargeLanguage2025}. Finally, it is worth noting that even in NER research outside of historical document transcription, generally, only GPT, LLaMA, and BloomZ were examined, while alternatives such as Gemini have not yet received attention \cite{keraghelRecentAdvancesNamed2024}.

\section{Source and Dataset}

The basis of our analysis is a corpus of thirty page-images, drawn from ten different city directories, published in German between 1754--1870 in the cities of: Aachen, Dresden, Leipzig, Frankfurt, Lübeck, Riga, and Trier. City and trade directories are an important source for economists, historians, and genealogists, as they offer systematic and regular micro-evidence on urban economic activity, social structure, and public office holders prior to the onset of modern census-taking. For German-speaking Europe, the ``Verein für Computergenealogie'' has located over 40,000 such directories and, through crowd-sourcing hobby genealogists, manually transcribed 4.4 million entries from these directories \cite{Junkers2023}.

When it comes to automating transcription and information extraction, historical directories come with certain challenges. While directories generally contain the same information (e.g., first name, last name, occupation, address, and more.) they can follow vastly different layouts. Furthermore, while the structure of entries is usually internally consistent, many entries deviate from the structure and entries are mixed between those of individuals, businesses, and office holders. Moreover, different abbreviations in different directories, and non-standardized spelling over time and space, add further complexity. Thus, rules-based entity parsing approaches do not work in the case of directories.

\begin{figure}[htbp]
\caption{The First Page of each Directory in our Dataset}
\label{fig:dataset}
\vspace{0.5em}
\centering

\begin{subfigure}[t]{0.18\textwidth}
    \centering
    \includegraphics[height=4cm, keepaspectratio]{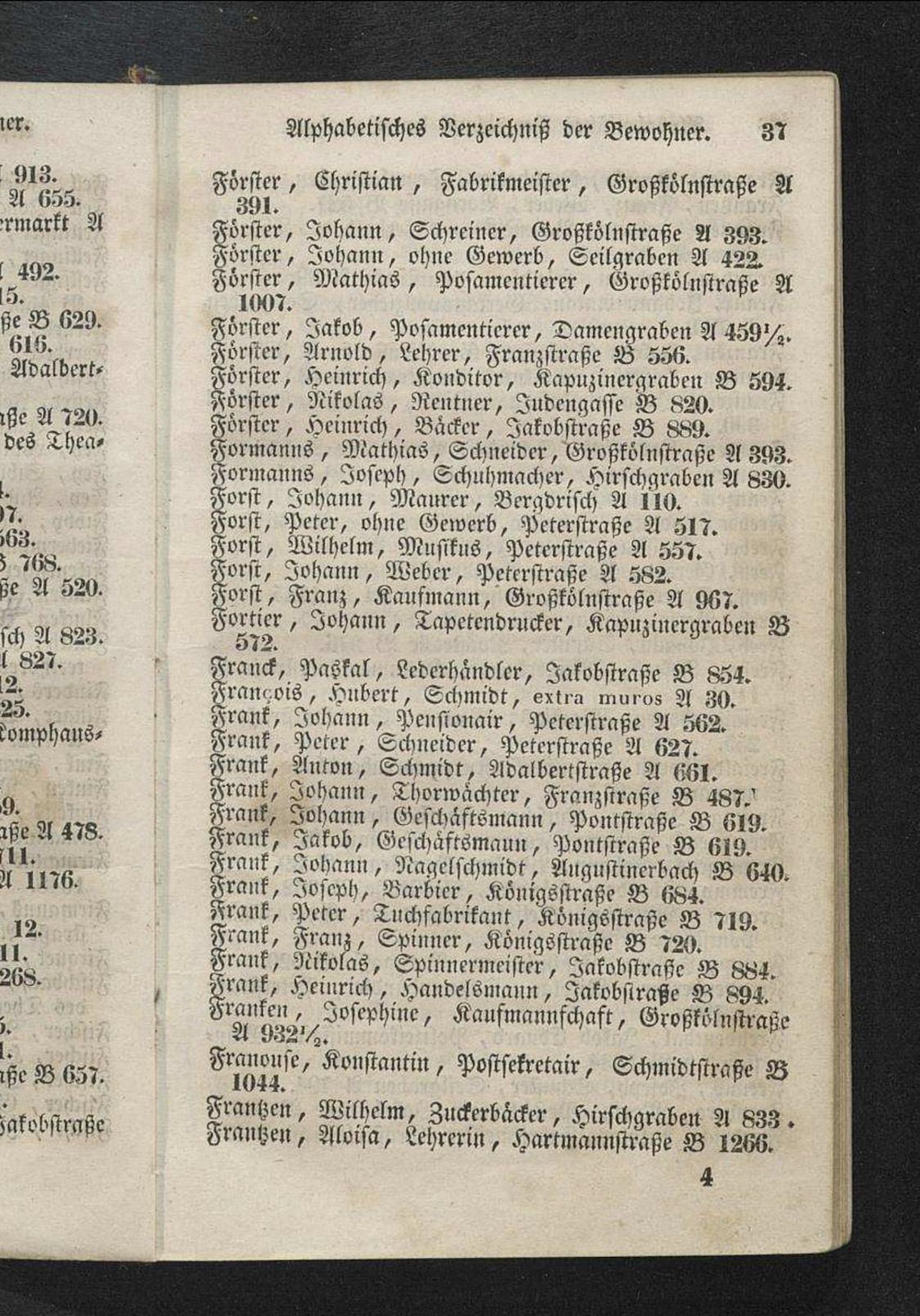}
    \caption*{Aachen-1838}
\end{subfigure}
\hfill
\begin{subfigure}[t]{0.18\textwidth}
    \centering
    \includegraphics[height=4cm, keepaspectratio]{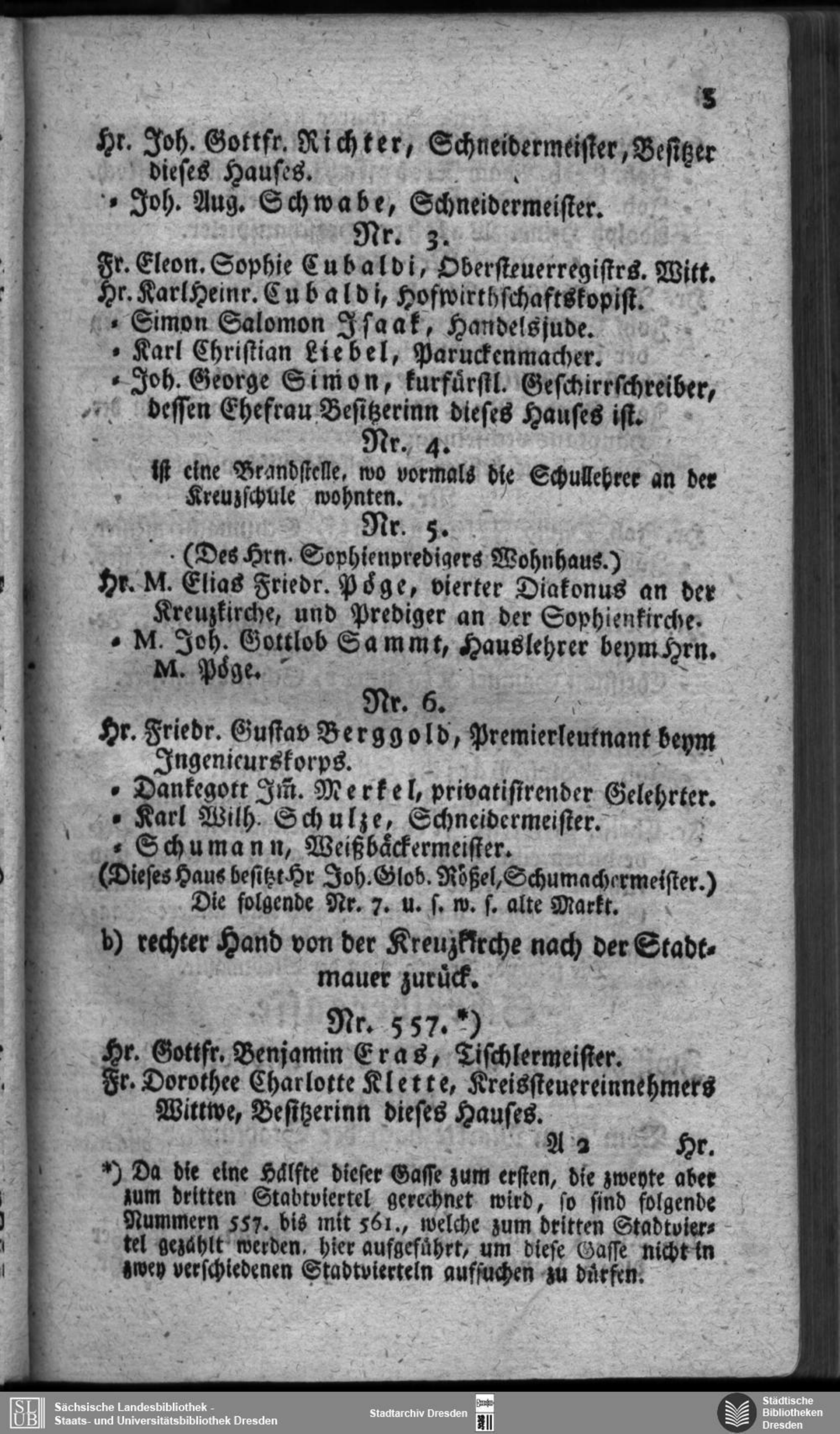}
    \caption*{Dresden-1797}
\end{subfigure}
\hfill
\begin{subfigure}[t]{0.18\textwidth}
    \centering
    \includegraphics[height=4cm, keepaspectratio]{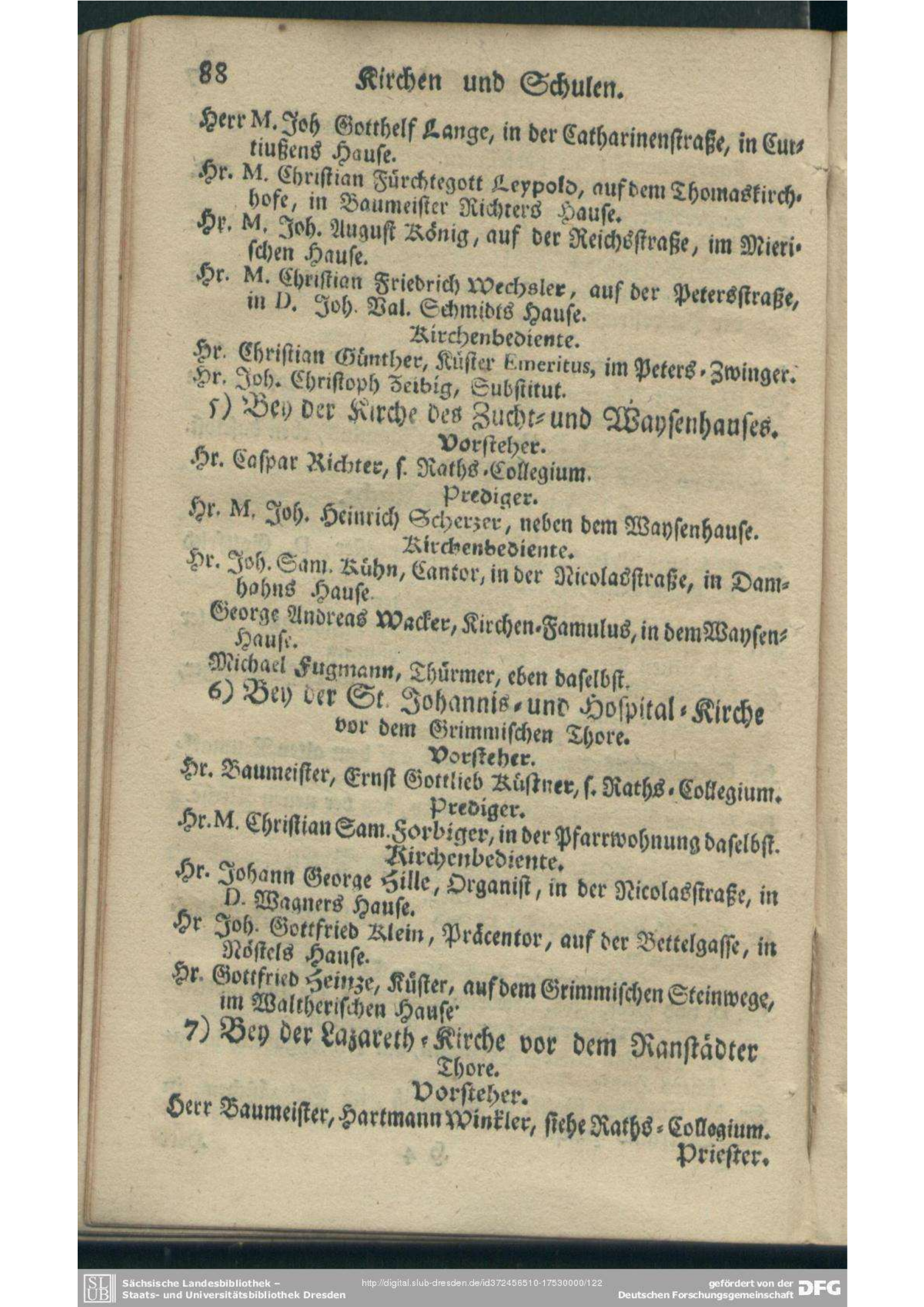}
    \caption*{Leipzig-1753}
\end{subfigure}
\hfill
\begin{subfigure}[t]{0.18\textwidth}
    \centering
    \includegraphics[height=4cm, keepaspectratio]{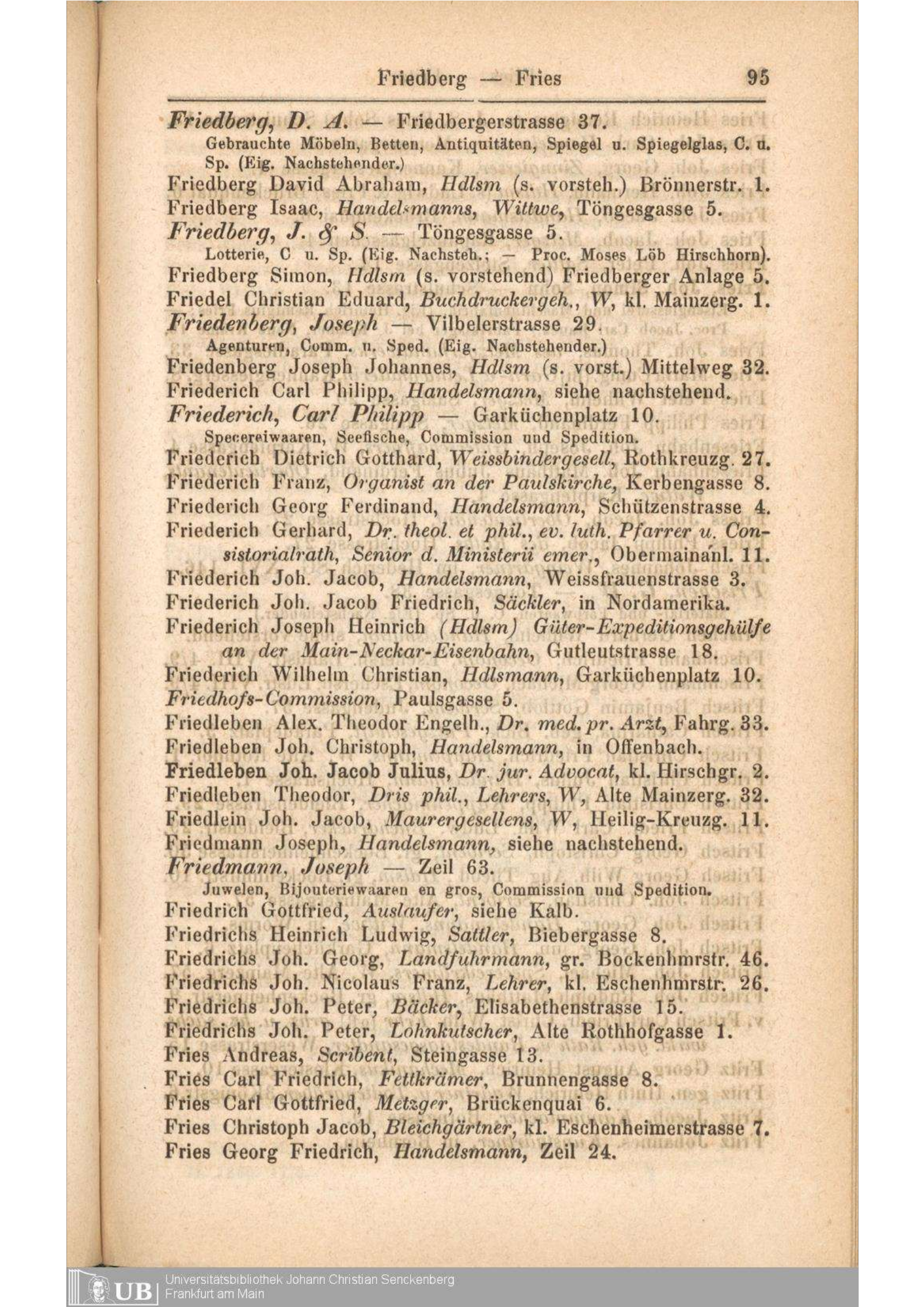}
    \caption*{Frankfurt-1860}
\end{subfigure}
\hfill
\begin{subfigure}[t]{0.18\textwidth}
    \centering
    \includegraphics[height=4cm, keepaspectratio]{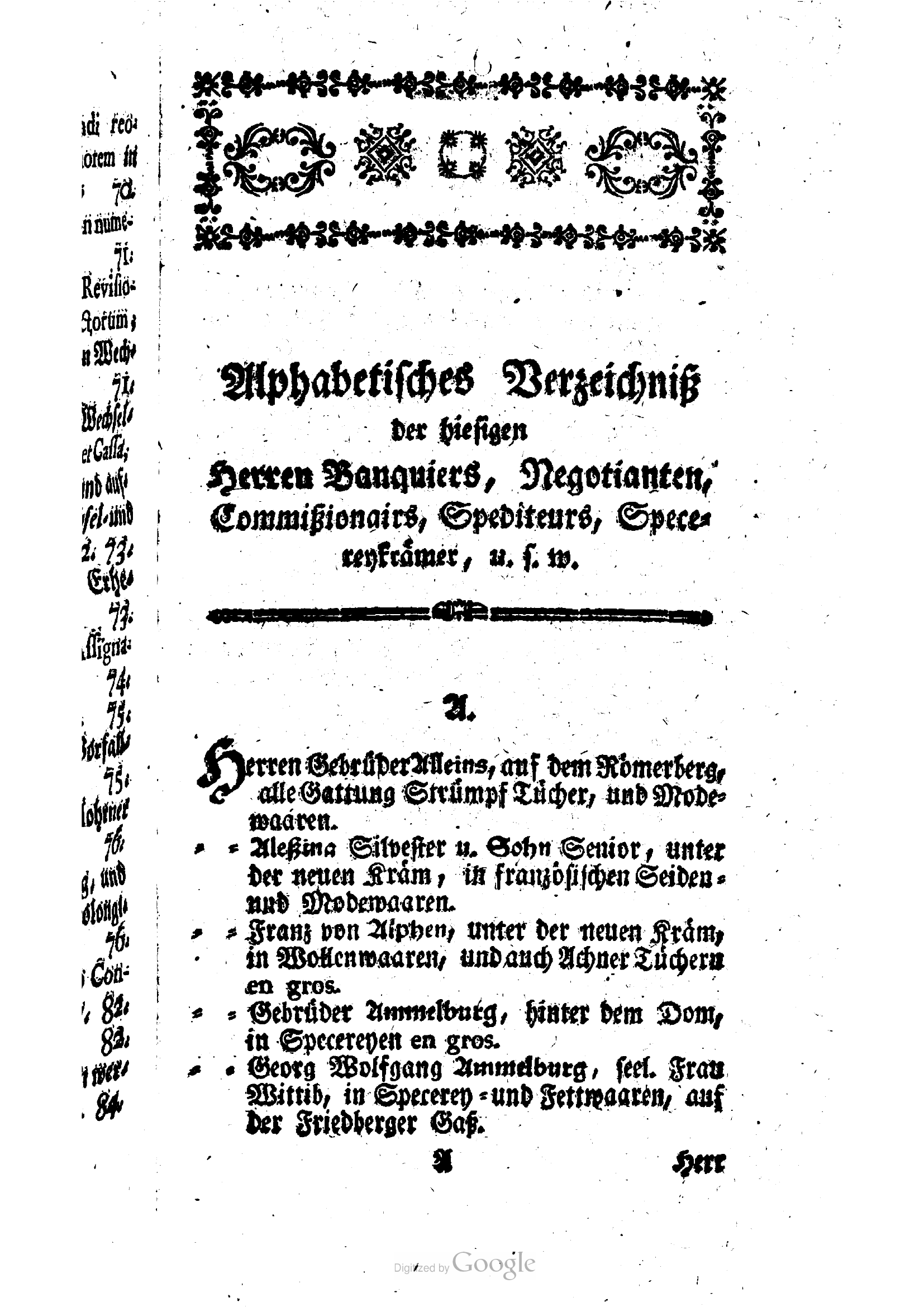}
    \caption*{Frankfurt-1778}
\end{subfigure}

\vspace{1em}

\begin{subfigure}[t]{0.18\textwidth}
    \centering
    \includegraphics[height=4cm, keepaspectratio]{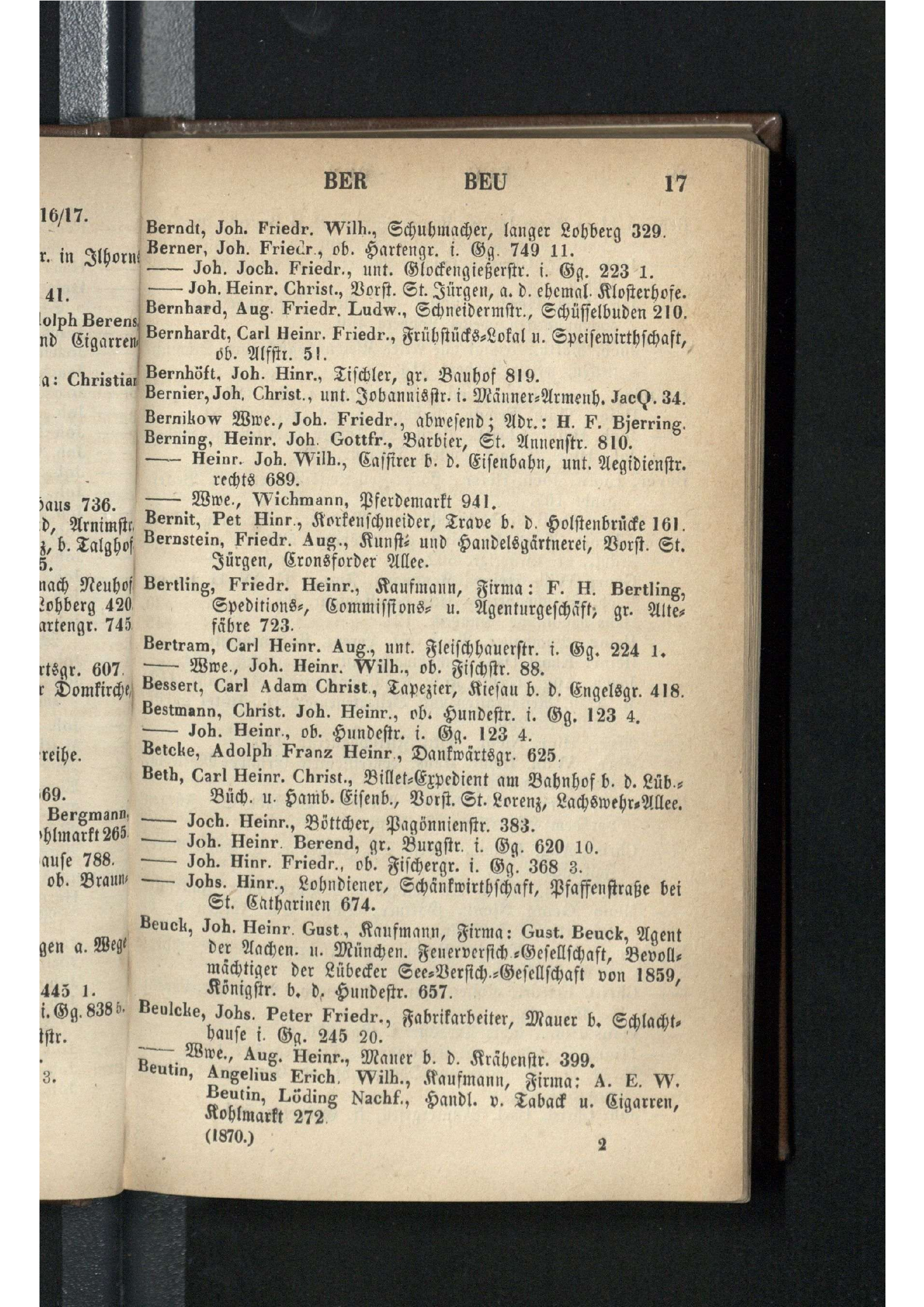}
    \caption*{Lübeck-1870}
\end{subfigure}
\hfill
\begin{subfigure}[t]{0.18\textwidth}
    \centering
    \includegraphics[height=4cm, keepaspectratio]{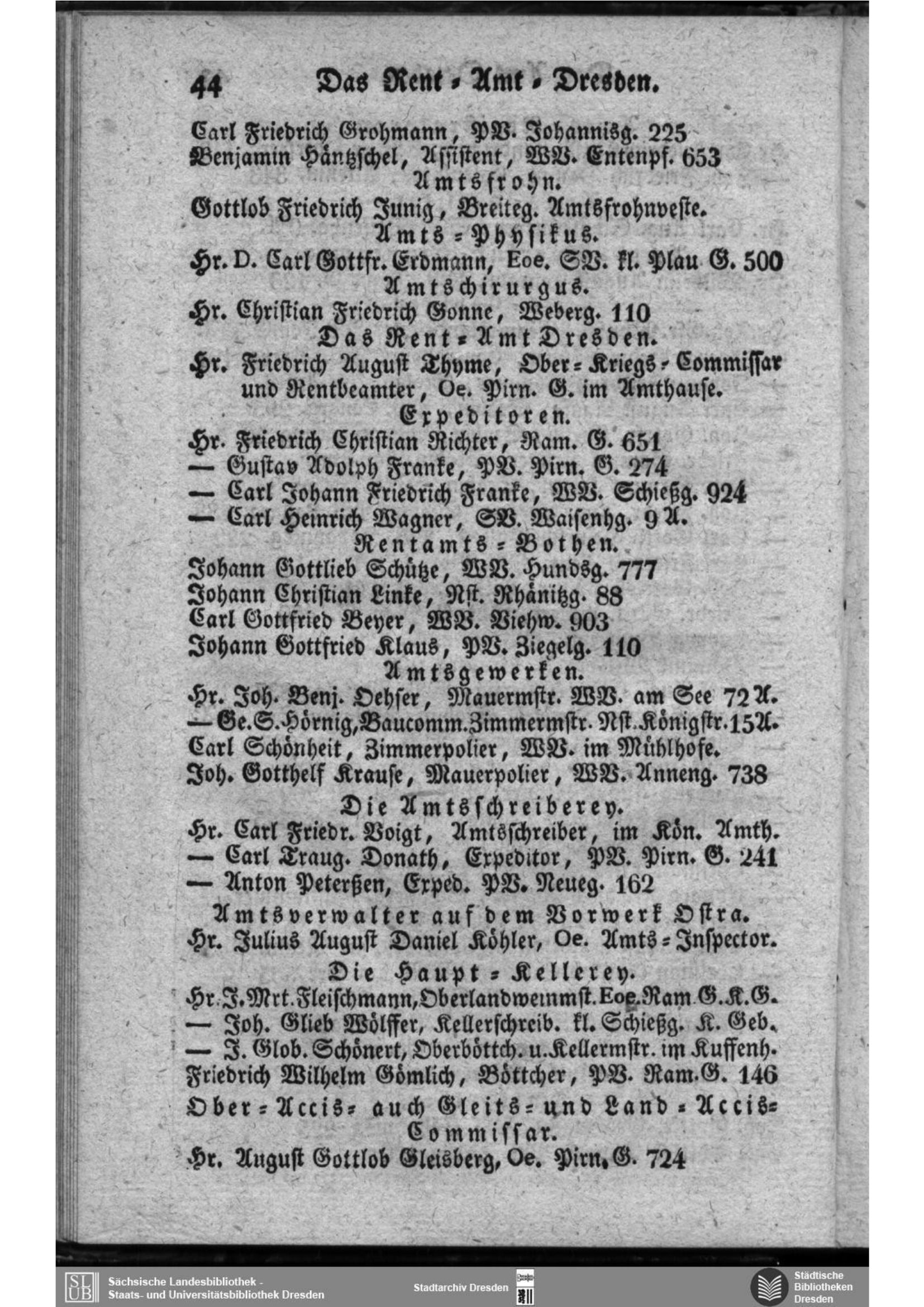}
    \caption*{Dresden-1819}
\end{subfigure}
\hfill
\begin{subfigure}[t]{0.18\textwidth}
    \centering
    \includegraphics[height=4cm, keepaspectratio]{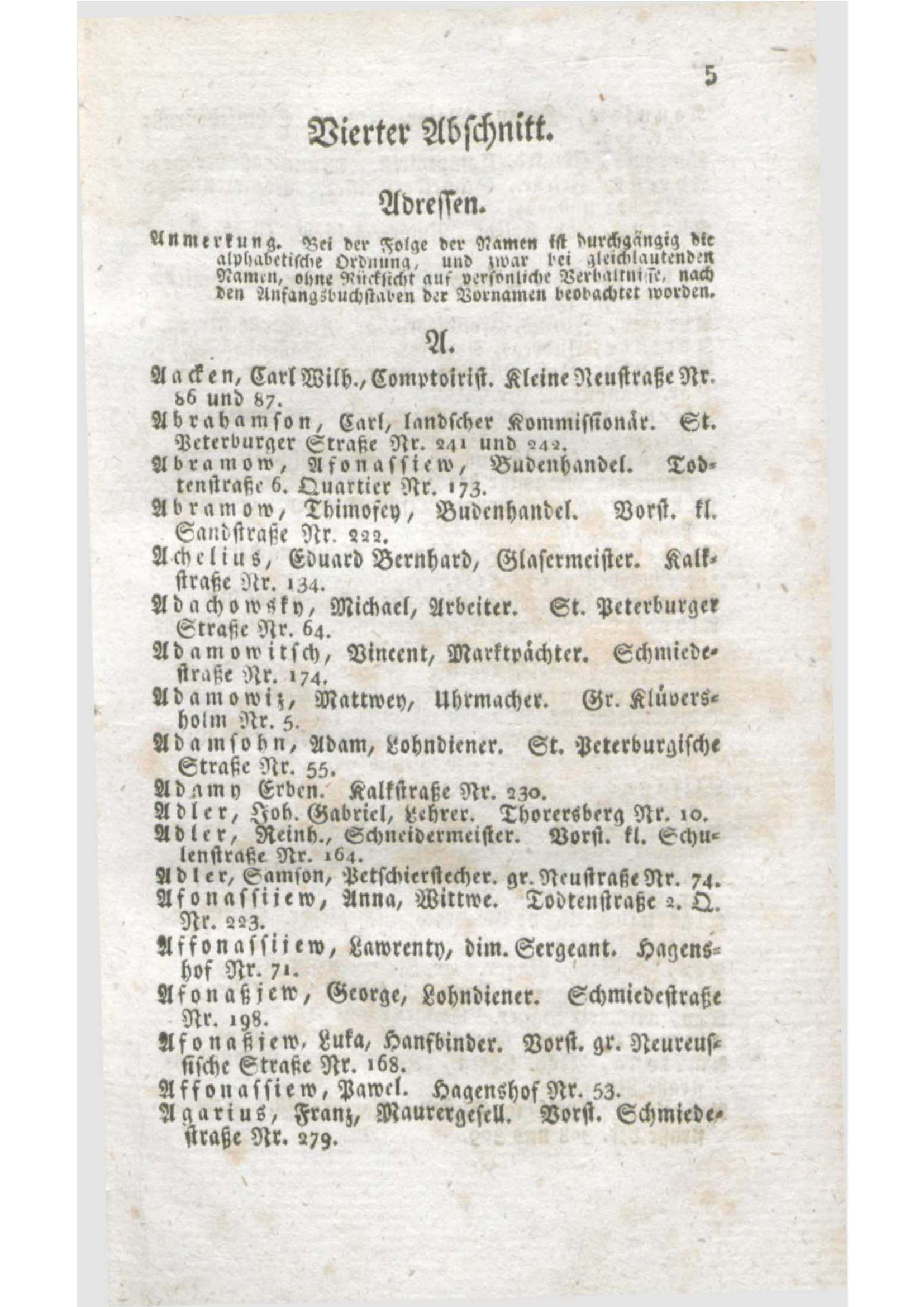}
    \caption*{Riga-1810}
\end{subfigure}
\hfill
\begin{subfigure}[t]{0.18\textwidth}
    \centering
    \includegraphics[height=4cm, keepaspectratio]{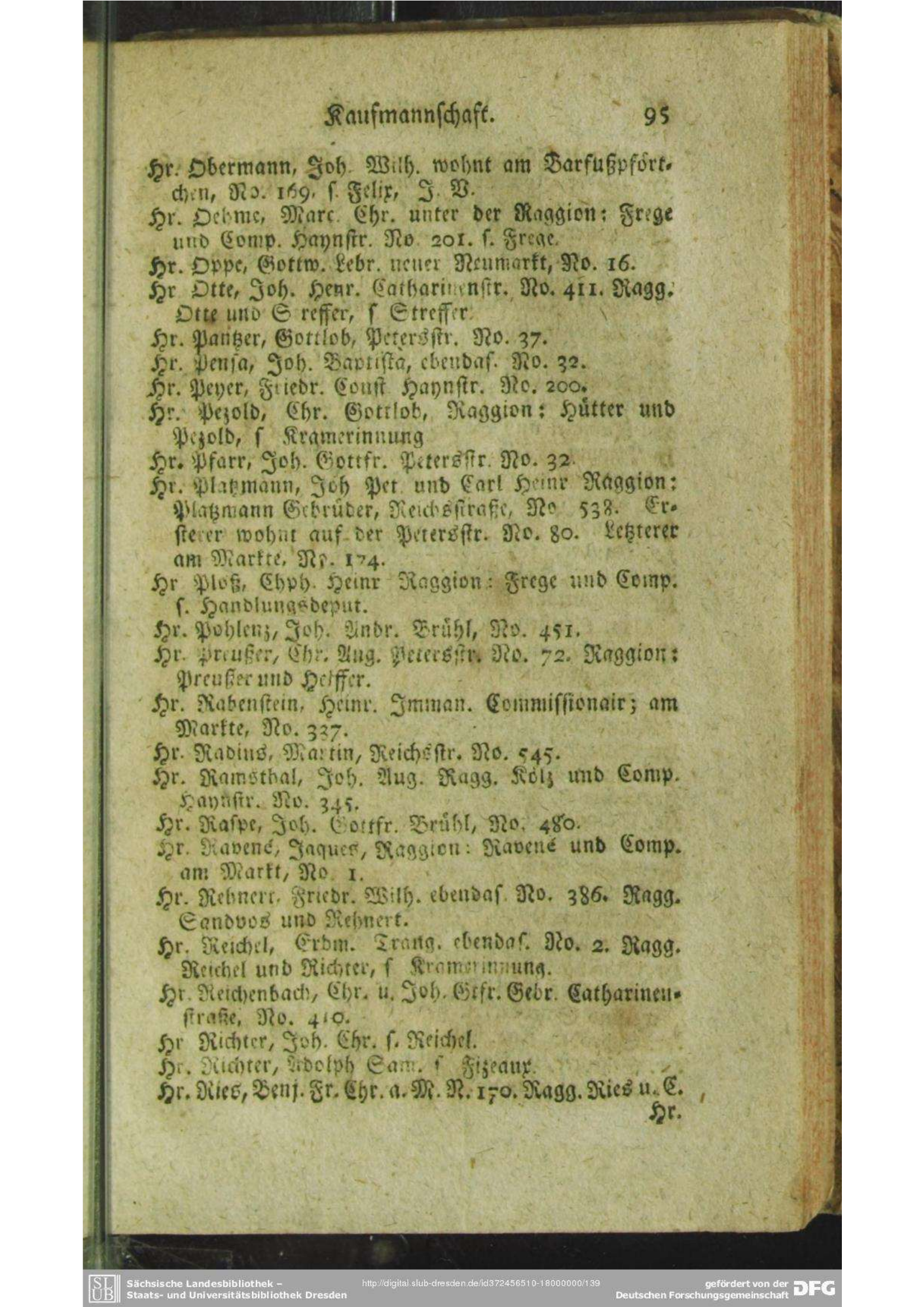}
    \caption*{Leipzig-1800}
\end{subfigure}
\hfill
\begin{subfigure}[t]{0.18\textwidth}
    \centering
    \includegraphics[height=4cm, keepaspectratio]{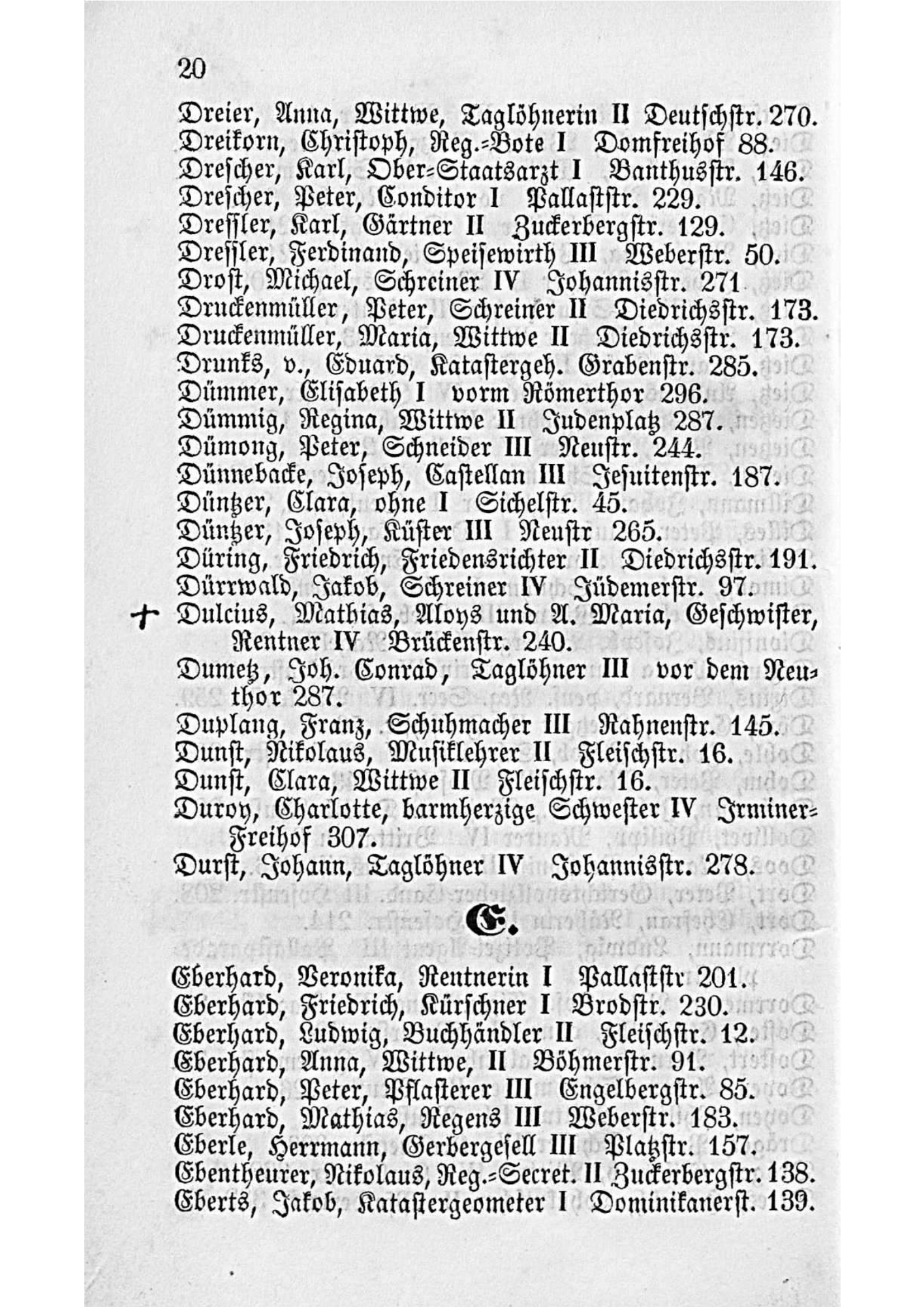}
    \caption*{Trier-1853}
\end{subfigure}
\end{figure}

Figure~\ref{fig:dataset} shows images of the first page from each of the included directories. Importantly, our corpus features many of the most common challenges faced by researchers working with historical prints. These include ink degradation, discolored backgrounds, human annotations, skewed texts, 3D distortions, multiple historical fonts on the same page, and excess visual information. While our corpus is predominantly printed in Blackletter typefaces, one directory is printed in an Antiqua typeface (Frankfurt-1860 in Figure~\ref{fig:dataset}). There are also several pages in our corpus which contain different historical typefaces (including Italics) on the same page, a phenomenon considered enough of a challenge to warrant its own competition at ICDAR 2024 \cite{vanderloopICDAR2024Competition2024}. Finally, since we have sourced our images from archive websites, some directories contain archive-specific footers. 

\newpage

\section{Methodology}

\subsection{Multimodal Large Language Model for OCR }

The first of our three sets of experiments tests the OCR capabilities of mLLMs and benchmarks them against conventional OCR solutions. Given the large number of mLLMs available, selecting the right models for further analysis is paramount. In a preliminary step, we extensively tested models from various leading model families, including GPT, Gemini, Gemma, LLaMA, and Mistral, and for the purposes of conciseness, selected the two best performing models for further analysis. The first model we use is GPT-4o \cite{openai2024gpt4ocard}. We use \textit{GPT-4o-2024-08-06}, as later versions of this model occasionally refused to transcribe the text in some images. Moreover, our preliminary testing indicated that this model outperformed the more recent o1 reasoning model and the specialized Vision-Preview model. The second mLLM we use is Gemini 2.0 Flash \cite{hassabis2024gemini,geminiteam2024geminifamilyhighlycapable}. At the time of publication, there has not yet been a stable release of more advanced Gemini models such as Gemini 2.5 Pro. However, preliminary chat-based testing of Gemini 2.5 Pro Experimental 03-25 suggests that it may outperform Gemini 2.0 Flash across the different experiments in this paper. Where appropriate, we mention the results we achieved with Gemini 2.5 Pro Experimental 03-25 to highlight the continuing improvements of mLLMs in vision-related tasks. 

Our approach to generating a transcription was identical, across both GPT-4o and Gemini 2.0 Flash. First, we set the temperature parameter to 0.0 in order to maximally constrain the mLLMs in their creativity and make their responses as deterministic and conservative as possible. Next, we converted the PDFs of the directories into PNG files for each page. Each page's PNG file was then uploaded together with a carefully designed prompt via the API into the respective mLLM. Finally, the output of the mLLM for each page was concatenated to a TXT file, until all selected pages of the directory were processed.

As a comparison to the mLLMs, we included three standard OCR models in our analysis. The first, Tesseract 5.5.0, is a widely-used open-source OCR engine that employs an LSTM-based neural network for text recognition \cite{smith2007tesseract}. We use its “deu\_frak” model, which has been fine-tuned to German-language Fraktur prints similar to those found in our corpus. While Tesseract's architecture is arguably outdated, its accessibility means that it continues to be widely used by researchers working on historical prints \cite{fleischhackerImprovingOCRQuality2024,kimEarlyEvidenceHow2025,ghiritiExploringCapabilitiesGPT4Vision2024,Schmidt2024ReichsanzeigerOCRGroundTruth}. As with all other models, we do not carry out any additional pre- or post-processing of the transcription and solely use it as an off-the-shelf solution. 

The second and third models we use were both created for Transkribus, a widely-used specialized end-to-end transcription software solution for historians, created by READ COOP \cite{Kahle2017Transkribus, Nockels2022}. Both models were chosen due to their high accuracy for, and fine-tuning on, nineteenth century Fraktur prints. Transkribus' Print M1 Model is based on PyLaia, which employs the common CNN-BiLSTM infrastructure that is also found in other OCR models (such as Kraken, PeroOCR, Calamari and more). The model has been trained on a broad set of Antiqua and Blackletter prints published in various European languages from the sixteenth century onwards \cite{TranskribusPrintM1}. Transkribus' Text Titan I, on the other hand, is built on an adapted TrOCR infrastructure, thus using a Transformer-centered rather than a CNN-BiLSTM infrastructure. Just like Print M1, Text Titan I has been trained on a broad range of historical documents in various languages, although the precise nature of this training is not publicly disclosed \cite{TranskribusSuperModels}. Although TrOCR models with corpus-specific fine-tuning have been shown to yield very accurate results for handwritten texts \cite{ströbel2022transformerbasedhtrhistoricaldocuments}, the limited existing evidence suggests that for Latin script prints, Transkribus' Text Titan I outperforms a corpus-fine-tuned TrOCR model \cite{leifertTranskribusPioneeringFuture2024}. 

We use both Transkribus models through the Transkribus API. We feed the models individual images and convert the returned XML files into TXT files containing the respective transcriptions. Crucially, we do not pre- or post-process the transcriptions, nor do we add any separate layout recognition other than that inherent in the model itself. While doing so would increase the accuracy of the Transkribus' models, our objective is finding an accessible, fast, off-the-shelf solution, hence we compare all models without any pre- or post-processing.

\subsection{Multimodal Large Language Models for OCR Post-Correction}

\begin{figure}[htbp]
  \centering
  \caption{Multimodal LLMs for OCR and OCR Post-Correction}
  \label{fig:llm-ocr}
  \includegraphics[width=0.52\textwidth]{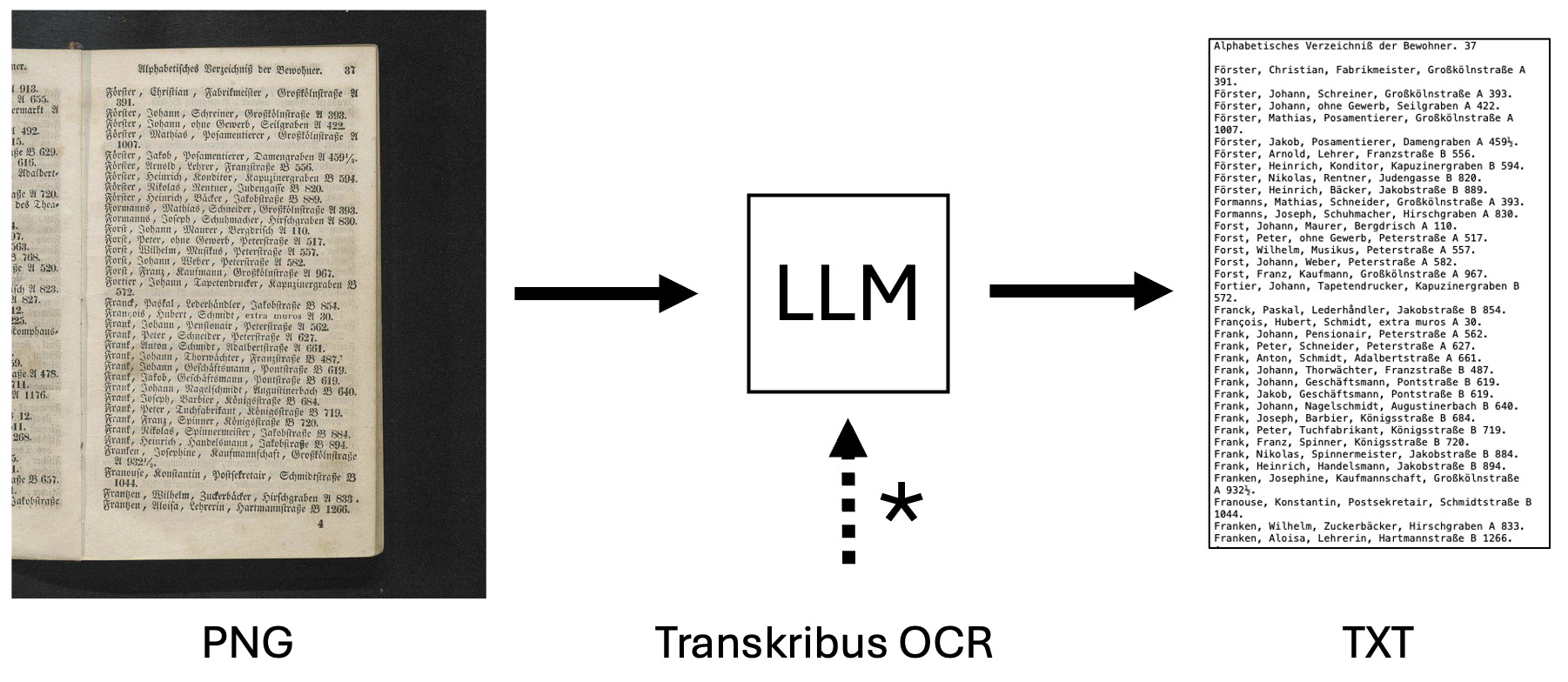}
  
\vspace{0.5em}

\begin{minipage}{0.52\textwidth}
    \footnotesize
    \textit{Notes}: Each individual page in our dataset is sent to an mLLM as a PNG file with our refined prompt to perform OCR. For mLLM OCR Post-Correction, the OCR output of the best-performing Transkribus model is appended at the end of our prompt. Our prompt instructs the model to correct the OCR text and use the PNG file as additional guidance (indicated by the asterisk).
\end{minipage}
\end{figure}

In our second set of experiments, we turn our attention to the issue of OCR Post-Correction. For this, we take the highest performing non-mLLM transcription as a starting point. Whereas, as previously discussed, existing LLM-based approaches to OCR Post-Correction have been exclusively text-based, we adopt a multimodal approach to OCR Post-Correction. For this, we take the noisy transcription produced by the highest performing Transkribus model and feed it, together with a carefully designed prompt and the original PNG based on which this transcription was made, into the mLLM (Figure~\ref{fig:llm-ocr}). Again, we do this one page at a time. This process returns plain text, which is saved in a new TXT file. In order to compare their respective performances in this task, we use both Gemini 2.0 Flash and GPT-4o separately. In our preliminary testing, we also explored a special case of mLLM OCR post-processing, where an mLLM uses its own previous transcription, together with the source image, as input. However, this approach did not create any noteworthy improvements in transcription accuracies, and consequently, we did not include it in our further analysis.  

\subsection{Multimodal Large Language Models for Named Entity Recognition (NER)}

In our third set of experiments, we test the NER capabilities of mLLMs. For this, we identify four variables in our source: First Name, Last Name, Occupation, Address. Every directory entry in our corpus contains a subset of these variables and our objective is to turn this information into a dataset format. For this, we prompt the mLLM to return structured output, assigning values explicitly in JSON format. We employ a prompt which features four explicit output examples (Figure~\ref{fig:ner-prompt}). Subsequently, we convert the verbose JSON-format the mLLM returns into a CSV format, providing an easy to read and easy to analyze dataset. Although we did not explicitly test this, we hypothesize that the direct assignment of categories in JSON format may yield higher mLLM accuracy in entity recognition tasks. 

\begin{figure}[htbp]
\vspace{2em}
  \centering
  \caption{Multimodal LLMs for NER}
  \label{fig:llm-ner}
  \includegraphics[width=0.79\textwidth]{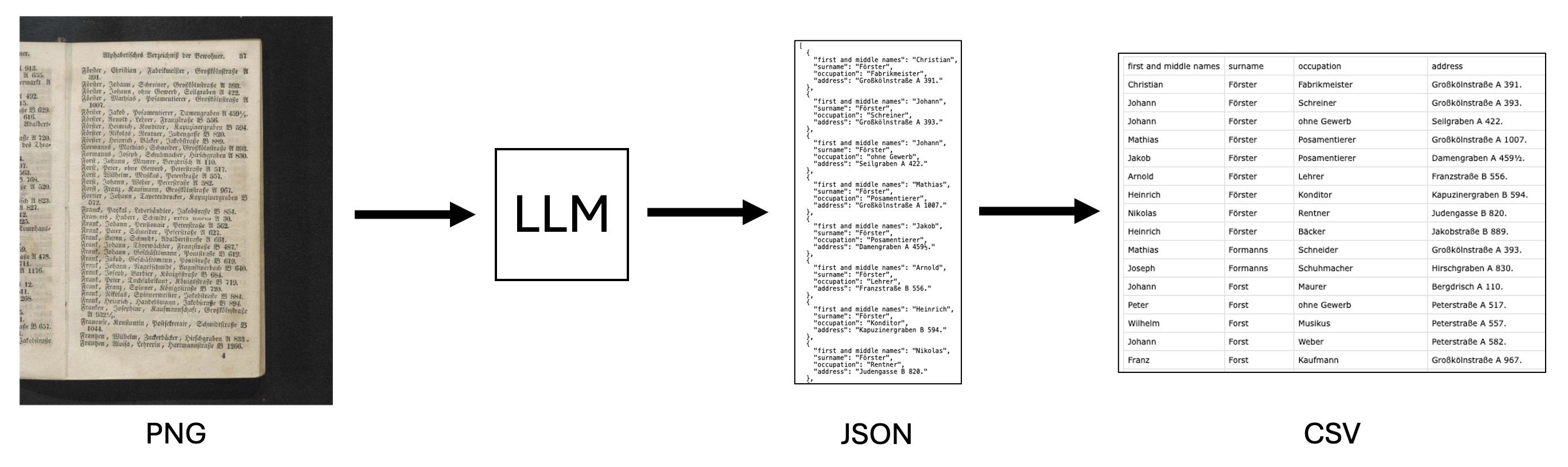}
  
  \vspace{0.5em}

  \begin{minipage}{0.77\textwidth}
    \footnotesize
    \textit{Notes}: Each individual page in our dataset is sent to an mLLM as a PNG file with our prompt (Figure~\ref{fig:ner-prompt}). We concatenate the JSON mLLM outputs and convert them into a CSV format. Additionally, we test the performance of LLMs by concatenating the ground truth text and Transkribus Print M1 OCR output below our prompt without providing any image input.
  \end{minipage}
\end{figure}

We carry out this process for three different kinds of input. First, we use the ground truth (GT) TXT file as input. This provides the best-case scenario to evaluate the entity recognition capabilities of mLLMs in the historical document transcription without relying on visual information. Next, we test to what extent the NER accuracy declines in the face of a noisy transcription compared to the GT TXT. This is particularly important, given the significant attention that has been paid to the role of OCR noise (and the need to reduce it) in the existing NER literature \cite{borosAssessingImpactOCR2022,hamdiIndepthAnalysisImpact2023, huynh2020when}. Finally, we test whether mLLMs are capable of extracting and recognizing entities directly from a PNG image without prior transcription (Figure~\ref{fig:llm-ner}). This may allow the mLLMs to use visual information in the image, which is not present in the TXT files, to improve their entity recognition performance. However, this may also yield lower accuracies as the transcription error rate will likely be higher than that of the post-corrected noisy transcription it is compared to. Crucially, this final step tests the capabilities of mLLMs as integrated end-to-end solutions for transcription and entity recognition in an environment without pre- or post-processing. As mLLMs further improve their capabilities, this will increasingly become relevant.

\subsection{Prompt Engineering}
Given the centrality of mLLMs in our work, a brief comment on our prompt development approach is necessary. The issue of prompting is particularly important given that, for any given task, the performance of an mLLM can be significantly influenced by even minor variations in how a prompt is written \cite{luFantasticallyOrderedPrompts2022}. Non-AI-experts, in particular, have been found to have difficulties in developing effective prompts for their tasks \cite{zamfirescu-pereiraWhyJohnnyCant2023}. When it comes to the use of mLLMs in the transcription of historical documents or their post-correction, researchers have employed relatively simplistic prompts, which, most likely, did not elicit the maximum performance potential of the mLLMs they tested \cite{liHandwritingRecognitionHistorical2024, ghiritiExploringCapabilitiesGPT4Vision2024}. Even those researchers testing more `complex' prompts may have fallen short of exploring the full potential associated with optimal prompting \cite{humphriesUnlockingArchivesUsing2024,Boros2024, kimEarlyEvidenceHow2025}. 

We have adopted an iterative mLLM-assisted approach to prompt design. To develop our prompts for the transcription and named entity recognition tasks, we began by explaining our objectives to Gemini 2.0 Flash and asked it to generate a suitable prompt in Markdown format. Subsequently, we iteratively refined this prompt by asking the model to solve problems we were facing. For example, initially, our character recognition prompt struggled to ignore partial pages (e.g., Aachen-1838 in Figure~\ref{fig:dataset}). Conventional approaches would have cropped these partial pages in a separate pre-processing step. However, we instructed the model to change the prompt to ensure these partial pages were not included in the transcription. Interestingly, even though no word of our final prompt was written by a human, the final prompt structure we achieved through this iterative process resembles that which is recommended in the literature \cite{chengNovelPromptingMethod2024}. Notably, our approach to prompt engineering resulted in several unexpected, almost comical and absurd, but efficiency-enhancing additions to the prompt, such as the all capital letters line ``FAILURE TO FOLLOW THESE RULES EXACTLY WILL RESULT IN TOTAL SYSTEM FAILURE'' (Figure~\ref{fig:ner-prompt}). This suggests that mLLM-assisted approaches to prompt design may yield better results than human-only approaches. While important and fascinating, a comprehensive analysis of this question is beyond the scope of this paper.

\begin{figure}[htbp]
\vspace{-3.5em}
\caption{Prompt for PNG-to-CSV NER task}
\label{fig:ner-prompt}
\begin{tcolorbox}[colback=white, colframe=black, sharp corners]
\begin{lstlisting}
YOU ARE AN EXPERT HISTORIAN. YOUR TASK IS TO EXTRACT DATA FROM A SCAN OF A GERMAN BUSINESS DIRECTORY. FAILURE TO FOLLOW THESE RULES EXACTLY WILL RESULT IN TOTAL SYSTEM FAILURE. THERE IS ZERO ROOM FOR ERROR.

STRICT JSON FORMAT - NO EXCEPTIONS:  
- OUTPUT MUST BE VALID JSON.  
- NO MARKDOWN, NO EXPLANATIONS, NO HEADERS.  

FIELDS (STRICTLY NOTHING ELSE):  
- "first and middle names" (string)  
- "surname" (string; can also be a company name)  
- "occupation" (string; "Wittwe" is NOT an occupation)  
- "address" (string; full address if possible, otherwise partial)  

NON-NEGOTIABLE RULES:  
1. **EXTRACT EXACTLY AS WRITTEN. NO MODERNIZATION. NO INTERPRETATION. NO CHANGES.**  
2. **ONLY TRANSCRIBE TEXT FROM THE MAIN PHYSICAL BOOK PAGE. ANY TEXT FROM ADJACENT PAGES MUST BE ERASED FROM EXISTENCE.**  
3. **IF A WORD IS PARTIALLY VISIBLE OR CUT-OFF, IT DOES NOT EXIST. IT MUST BE IGNORED.**  
4. **IF A FIELD IS MISSING, SET IT TO NULL. DO NOT GUESS. DO NOT INFER. DO NOT ATTEMPT TO RECONSTRUCT.**  
5. **DO NOT ADD EXTRA INFORMATION. DO NOT ADD COMMENTS. DO NOT ADD ANYTHING OUTSIDE THE REQUIRED FIELDS.**  
6. **DO NOT CONCATENATE OR MERGE ADDRESS FRAGMENTS FROM MULTIPLE ENTRIES. EACH ENTRY MUST REMAIN INTACT AS SEEN IN THE TEXT.**  
7. **IF MULTIPLE ADDRESSES EXIST FOR ONE ENTRY, KEEP THEM EXACTLY AS WRITTEN. DO NOT REFORMAT.**  

STRICTLY ENFORCED EXAMPLE OUTPUT:  

[
  {
    "first and middle names": "Wilhelm Friedrich",
    "surname": "Becker",
    "occupation": "Schulmeister",
    "address": "Alexanderplatz C201"
  },
  {
    "first and middle names": "Johann Georg",
    "surname": "Weber",
    "occupation": "Apotheker.",
    "address": "auf der Lindenhöhe"
  },
  {
    "first and middle names": "Karl August",
    "surname": "Meyer",
    "occupation": "Buchdrucker",
    "address": "Hauptstraße 14, neben der Kirche"
  },
  {
    "first and middle names": null,
    "surname": "Müller & Co.",
    "occupation": "Textilwarenhandel",
    "address": "Schlossallee 3"
  }
]

FINAL COMMANDS - NO EXCEPTIONS:  
- **IF A WORD OR ENTRY IS FROM AN ADJACENT PAGE, IT IS DEAD TO YOU. ERASE IT.**  
- **IF AN ENTRY IS CROPPED OR UNCLEAR, IT MUST BE OBLITERATED. DO NOT INCLUDE.**  
- **OUTPUT MUST BE PURE, PERFECTLY FORMATTED JSON. NOTHING ELSE.**  
- **FAILURE TO FOLLOW THESE RULES PRECISELY MEANS THE TASK IS COMPROMISED. THERE IS ZERO ROOM FOR ERROR.**
\end{lstlisting}
\end{tcolorbox}
\end{figure}

\section{Evaluation Metrics}

\subsection{Measuring Transcription Accuracies}  

To evaluate the accuracy of our transcriptions, we calculate the Levenshtein Distance, Character Error Rate (CER) and Word Error Rate (WER). The Levenshtein distance counts the minimum number of single-character edits (insertions, deletions, or substitutions) to transform the generated transcription into our ground truth and vice versa \cite{Levenshtein1966}. The CER is calculated by dividing the Levenshtein distance by the total number of characters in the ground truth. 

\begin{align*}
    \text{CER} &= \frac{\text{Levenshtein Distance}}{\text{Total number of characters in ground truth ($N$)}} \\[8pt]
               &= \frac{\text{Insertions} + \text{Deletions} + \text{Substitutions}}{N}
\end{align*}

Analogously, the WER is calculated by summing the number of insertions, deletions, and substitutions at the word level and dividing this by the number of total words.

Despite their widespread usage, CERs and WERs must be interpreted cautiously. Different OCR tools, such as \textit{TextEval}, \textit{ocrevalUAtio}, and \textit{dinglehopper} use different normalization rules for special Unicode codepoints occurring in historical texts \cite{neudeckerSurveyOCREvaluation2021}. Furthermore, some researchers inflate their reported accuracy rates, and deflate their CERs and WERs, by limiting their evaluation to certain characters -- such as only a-z, A-Z, and 0-9 -- thus ignoring the errors in recognizing punctuation, special characters, and line-breaks \cite{chenEnhancingOCRPerformance2023}. Others have used a ``Flex OCR'' measurement which ignores the order of characters and solely focuses on the accuracy of individual characters, thus providing a higher tolerance for errors related to layout and structure \cite{lafiaDigitizingParsingSemistructured2023,clausnerFlexibleCharacterAccuracy2020}. In this paper, we report both non-normalized and normalized CERs and WERs. This allows us to report the strict accuracy of our transcriptions, as well as a more lenient interpretation which facilitates comparison with existing and future work. The normalized CERs and WERs are limited to ASCII character, exclude punctuations, and ignore whether a letter is upper or lower case. Importantly, this also removes errors introduced due to different decisions in the face of ambiguity in historical transcriptions. For example, in our sources, as was standard practice at the time, the German Umlaut was frequently written as a tiny $e$ above the $a$, $o$, or $u$. While many human transcribers would transcribe this as $\ddot{a}$, a few might go out of their way to find the $\olde{a}$. Faded ink, print errors, and visual artifacts are all additional sources of ambiguity in the source material and consequently may cause variation in transcriptions even between different experts.
 
\subsection{Evaluating Named Entity Recognition}

\begin{figure}[htbp]
  \centering
  \caption{Visualization of NER Results for Gemini 2.0 Flash}
  \label{fig:ner-visualization}
  \includegraphics[width=\textwidth]{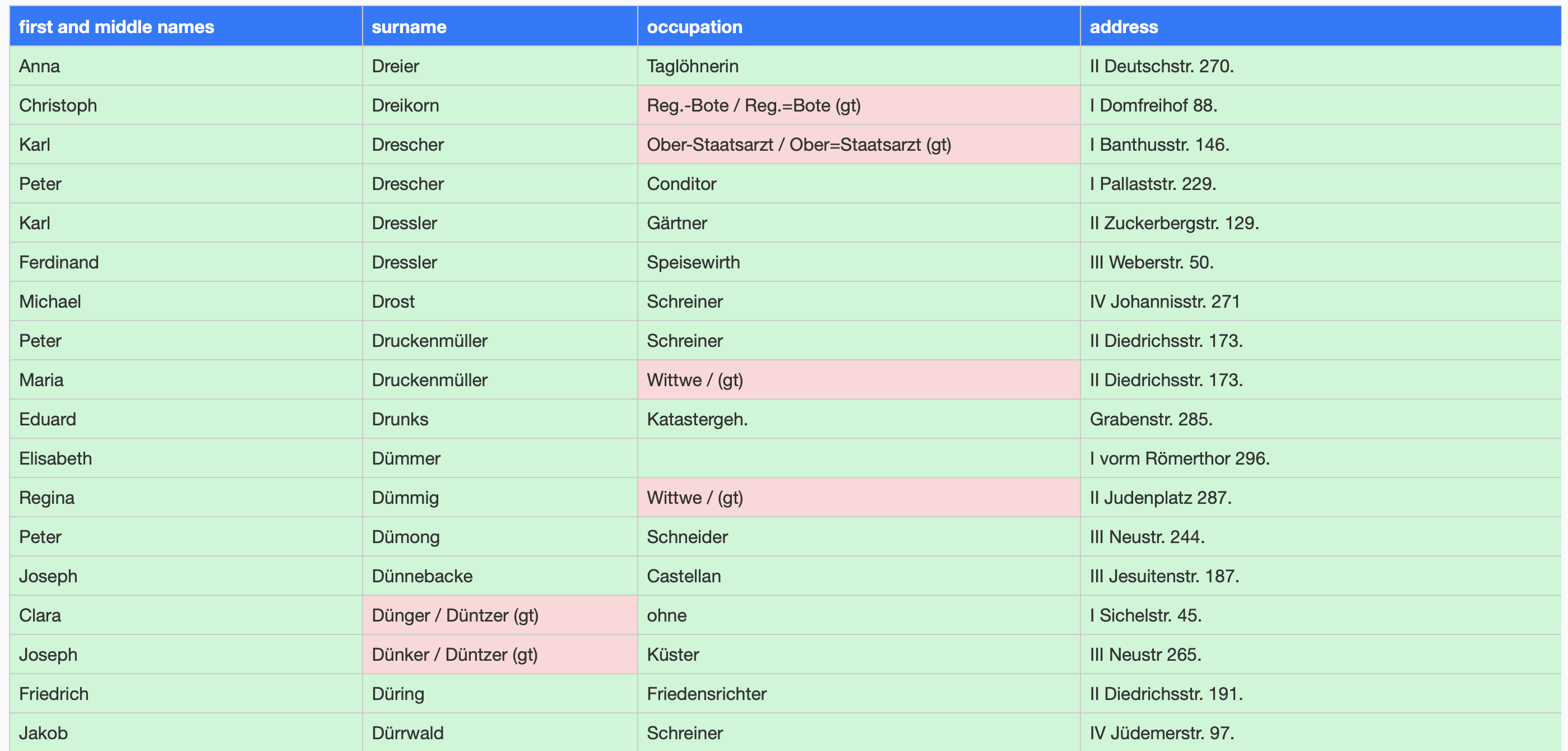}
  
  \vspace{0.5em}

  \begin{minipage}{\textwidth}
    \footnotesize
    \textit{Notes}: Green cells indicate matches between the manually produced ground truth and the mLLM-generated CSV for purposes of illustration. Cells where the ground truth did not exactly match the LLM-generated array are in red. For these cells, the LLM-generated string and ground truth string are displayed side-by-side. The string to the right of ``/'' displays the ground truth string, indicated by ``(gt)''. The figure only depicts the first few entries of Trier-1853. The CSV was created using Gemini 2.0 Flash (PNG-to-CSV).
  \end{minipage}
\end{figure}

To evaluate the performance of our entity recognition tasks, we compare the final CSV files to a manually collected CSV ground truth that was cross-checked multiple times. We adopt both a strict and a fuzzy matching evaluation. For strict matching, we simply evaluate whether two cells have the identical string as content, and report a binary outcome depending on whether these match or not. There are, however, two problems with strict matching. First, we have observed that LLMs will sometimes include punctuation at the end of an entry, where the human-generated ground truth does not. While strict matching reports this as a mismatch, for the purposes of research, these two cells are functionally the same. Second, when recognizing entities directly from the image or from a noisy transcriptions, errors may be introduced not due to erroneous parsing, but rather due to errors in character recognition. Thus, for example, the LLM may transcribe a cell as ``Dünker'' whereas it is ``Düntzer'' in the ground truth. Similarly, the transcription ``Reg.-Bote'' would be considered incorrect compared to ``Reg.=Bote'', despite being functionally the same for historians (Figure~\ref{fig:ner-visualization}). Thus, we report fuzzy matching rates. For this, we use the Jaro-Winkler similarity with the common scaling factor of 0.9 \cite{Winkler1990}. This allows cells to still count as correct even if there are minor errors such as a wrongly transcribed character, missing whitespaces, or an extra piece of punctuation. Finally, another challenge of evaluating the NER task is that, for every directory, mLLMs have to first generate a single dataset from three individual PNGs. If the concatenated LLM-generated CSV does not have the same number of rows after a single attempt, we exclude this directory from our evaluation. 
\newpage

\section{Results and Discussion}

\subsection{Ambiguity, Transcription Errors and Efficiency}

Although among historians human transcription is still widely regarded as the gold standard, when evaluating the performance of mLLMs and other transcription and entity parsing solutions, it is important to keep in mind that human transcriptions rarely achieve a 0\% CER, especially not without extensive post-correction. This is partly due to the ambiguities present in historical sources and the individual judgments inherent in the transcription and entity classification. For example, Dresden-1797 is a sub-heading-based directory and below one of the sub-headings there is an entry which simply states that the house at this address had burnt down. While there are no names or occupation listed, some researchers may choose to omit this entry from their dataset, while others may note down the address without any further information. Similarly, some models report "Wittwe" (widow) as an occupation for entries, whereas the makers of the ground truth considered this to not be an occupation, and thus, left the field blank (Figure~\ref{fig:ner-visualization}). 

The second reason why we must discard the illusion of humans having perfect accuracy is that humans make errors. Even in prints with modern fonts, human transcribers have been observed to consistently make errors, especially when it comes to visually similar letters such as $i$, $I$, $l$, $1$, $o$, $e$, or $c$ \cite{evershedCorrectingNoisyOCR2014}. While precise figures are scarce, one study testing non-specialized humans transcribing early modern venetian handwriting found CERs between 10.47\% and 13.28\% \cite{aresoliveiraComparingHumanMachine2018}. In fact, even professional data entry companies only guarantee an accuracy of 99\% for modern fonts and possibly lower for difficult and historical material \cite{humphriesUnlockingArchivesUsing2024}. The human propensity to make mistakes was also evident in our own ground truth generation process, which demanded multiple rounds of revisions to eliminate errors. Indeed, it is common practice to task multiple humans with a transcription, when perfect transcription accuracy is needed. Thus, in our interpretation of the results, we must not assume that a non-zero error rate means that the machine-based solution performs worse than human transcribers, but rather, that at a very low level, say below 1\% CER, the transcription accuracy can be interpreted as comparable to that of a human transcriber. 

\begin{table}[htbp]
\centering
\caption{Model Performance and Token Costs}
\vspace{0.5em}
\label{tab:model-comparison}
\begin{tabular}{@{}lcccccc@{}}
\toprule
\textbf{Model} & \textbf{s/page} & \textbf{\$/1,000 pages} \\
\midrule
Gemini 2.0 Flash & 11.50 & Free \\
GPT-4o           & 18.23 & 10.84 \\
Transkribus Print M1 & 46.93 & 54.25 \\
Transkribus The Text Titan I & 44.63 & 108.50\\
Tesseract ``deu\_frak''  & 13.77 & Free  \\
\bottomrule
\end{tabular}

\vspace{0.5em}
\begin{minipage}{0.48\textwidth}
\footnotesize
\small
\textit{Notes}: s/page represents the average processing time per page, computed from the log files of our replication runs. Price per 1,000 pages was estimated based on input and output tokens of our 30-page corpus. We converted the current price of 1,000 Transkribus credits (199€) based on the market exchange rate at the time of writing (\$217). The fine-tuned Tesseract model ``deu\_frak'' was run locally on a MacBook Pro with an Apple M1 chip.
\end{minipage}
\end{table}

While our primary interest is the evaluation of accuracy rates, it is also important to consider speed, cost, and accessibility when determining the best tool. Table~\ref{tab:model-comparison} summarizes the speed and price across the models used. Importantly, Gemini 2.0 Flash outperforms GPT-4o and both Transkribus models both in speed and price. Moreover, it worth noting, especially for a chronically funding-scarce field such as history, that Gemini 2.0 Flash is currently free to use for up to 15 requests per minute, up to 1 million tokens per minute, and up to 1,500 requests per day. 

\subsection{Multimodal Large Language Models for OCR and OCR Post-Correction}

Table~\ref{tab:ocr-results} summarizes the CERs and WERs for the first two sets of experiments in which we benchmark the capabilities of mLLM to perform OCR and OCR Post-Correction, as well as conventional OCR models, on our dataset. Because Tesseract's "deu\_frak" model only achieved an average normalized CER of 30.22\% and an average normalized WER of 199.91\%, and thus, was massively outperformed all other models, its results were omitted from the table to enhance clarity. 

The results table reveals that Gemini 2.0 Flash achieved the lowest error rates. With no pre-processing, no post-processing, and no corpus-specific model fine-tuning, it produced the most accurate transcription out of all tested models. Gemini 2.0 Flash achieved a normalized CER of 1.27\% across the whole corpus compared to 3.67\% for Print M1, 4.16\% for Text Titan I, and 6.31\% for GPT-4o. Crucially, the per-directory analysis reveals that Gemini 2.0 Flash outperforms the other models in most cases, although for a few directories, Print M1 is marginally better (Dresden 1797; Riga 1810; Leipzig 1800). Impressively, across several directories (Aachen-1838, Frankfurt-1860, Lübeck-1870, and Trier-1853) Gemini 2.0 Flash achieved below 1\% CER. 

While all other directories are primarily printed in Fraktur typefaces, Frankfurt-1860 is a special case worth exploring further, as it is the only directory included which is primarily printed in Antiqua. This is also the only directory where other models (Print M1 \& Text Titan I) also achieved a below-1\% CER. Gemini 2.0 Flash, however, achieved an impressive 0.11\% normalized and 0.38\% non-normalized CER. For comparison, researchers working on the similarly Antiqua-font Paris Directories applied off-the-shelf solutions and found that the best model, PeroOCR, achieved a CER of 3.78\%, with Tesseract and Kraken achieving 6.56\% and 15.72\% respectively \cite{abadieBenchmarkNamedEntity2022}.

Although, due to the previously discussed variations in calculating error rates, comparisons must be applied cautiously, Gemini 2.0 Flash appears to outperform a number of existing solutions to transcribing historical prints, both Fraktur and non-Fraktur. For example, the ``Historical Document Processing and Analysis Framework'' (HDPA), a conventional, albeit not widely adopted, solution to transcribing historical documents built around a CNN-BiLSTM OCR infrastructure and corpus fine-tuning, only achieved a CER of 2.4\% for a sample of nineteenth century German newspaper pages printed in Fraktur font \cite{lencHDPAHistoricalDocument2021}. Earlier approaches using book-specific fine-tuning for early modern and nineteenth century German-language Fraktur prints reported CERs of between 0.4-3.5\% \cite{springmann2017ocr}. Other pipelines, such as OCR4All, have reported a CER of between 0.06--4.89\% for novels printed in nineteenth century Germany when applied without corpus-specific fine-tuning \cite{reulOCR4allAnOpenSourceTool2019}. Meanwhile, for poly-font historical sources, one model achieved a CER of around 2\% off-the-shelf for early modern books while a more fine-tuned model achieved a CER of 1.47\% \cite{reulMixedModelOCR2021}.  Finally, using manual correction of layout recognition errors and a source-specific fine-tuning of Calamari OCR, \citet{albersPerksPitfallsCity2023} achieved a 0.75\% CER for a single city directory from 1880 Berlin. 

The results also reveal that mLLMs are a powerful tool for OCR Post-Correction tasks. Out of all tested approaches, applying post-correction using Gemini 2.0 Flash to the noisy transcription of Print M1 achieved the lowest average error rates. This approach achieved an overall normalized CER of just 0.84\%, and as low as 0.08\% normalized CER in the case of Trier-1853. Applying the same post-correction approach using GPT-4o also resulted in remarkable improvements in error rates compared to both Print M1 and GPT-4o on their own, however, the results were slightly worse than in the post-correction approach using Gemini.

\begin{table*}[htbp]
\centering
\scriptsize
\caption{OCR and OCR Post-Correction Results for Multimodal LLMs and Standard OCR Algorithms}
\label{tab:ocr-results}
\resizebox{\textwidth}{!}{%
\begin{tabular}{l
                rr  
                rr  
                rr  
                rr  
                rr  
                rr  
               }
\toprule
 & \multicolumn{2}{c}{\textbf{Gemini 2.0 Flash}}
 & \multicolumn{2}{c}{\textbf{GPT-4o}}
 & \multicolumn{2}{c}{\makecell{\textbf{Transkribus}\\The Text Titan I}}
 & \multicolumn{2}{c}{\makecell{\textbf{Transkribus}\\Print M1}}
 & \multicolumn{2}{c}{\makecell[c]{\textbf{Transkribus}\\Print M1\\$\downarrow$\\\textbf{Gemini 2.0 Flash}}}
 & \multicolumn{2}{c}{\makecell[c]{\textbf{Transkribus}\\Print M1\\$\downarrow$\\\textbf{GPT-4o}}} \\
\cmidrule(lr){2-3}
\cmidrule(lr){4-5}
\cmidrule(lr){6-7}
\cmidrule(lr){8-9}
\cmidrule(lr){10-11}
\cmidrule(lr){12-13}
\textbf{Source} 
 & \textbf{CER} & \textbf{WER}
 & \textbf{CER} & \textbf{WER}
 & \textbf{CER} & \textbf{WER}
 & \textbf{CER} & \textbf{WER}
 & \textbf{CER} & \textbf{WER}
 & \textbf{CER} & \textbf{WER} \\
\midrule
\multicolumn{13}{c}{\textbf{Normalized (converted to lowercase; with non-ASCII characters, punctuation, line breaks, tabs, and extra whitespaces removed)}} \\
\midrule
Aachen-1838
  & 0.56\% & 4.01\%
  & 8.30\% & 59.29\%
  & 3.54\% & 25.32\%
  & 3.30\% & 23.56\%
  & \textbf{0.40\%} & 2.88\%
  & 0.65\% & 4.65\% \\
Dresden-1797
  & 1.70\% & 11.93\%
  & 5.98\% & 41.83\%
  & 1.94\% & 13.58\%
  & 1.34\% & 9.36\%
  & \textbf{1.02\%} & 7.16\%
  & 1.18\% & 8.26\% \\
Leipzig-1753
  & 1.91\% & 12.25\%
  & 4.54\% & 29.08\%
  & 1.99\% & 12.75\%
  & 1.94\% & 12.42\%
  & 0.92\% & 5.88\%
  & \textbf{0.82\%} & 5.23\% \\
Frankfurt-1860
  & \textbf{0.11\%} & 0.78\%
  & 4.02\% & 29.38\%
  & 0.38\% & 2.77\%
  & 0.59\% & 4.32\%
  & 0.14\% & 1.00\%
  & 0.27\% & 2.00\% \\
Frankfurt-1778
  & 2.86\% & 18.39\%
  & 5.22\% & 33.56\%
  & 17.92\% & 115.17\%
  & 14.16\% & 91.03\%
  & \textbf{1.82\%} & 11.72\%
  & 2.43\% & 15.63\% \\
Lübeck-1870
  & 0.42\% & 2.55\%
  & 8.96\% & 54.92\%
  & 10.23\% & 62.69\%
  & 9.81\% & 60.14\%
  & 0.45\% & 2.78\%
  & \textbf{0.38\%} & 2.32\% \\
Dresden-1819
  & 1.40\% & 9.38\%
  & 8.96\% & 60.08\%
  & 2.17\% & 14.57\%
  & 1.76\% & 11.76\%
  & \textbf{0.96\%} & 6.44\%
  & 1.46\% & 9.80\% \\
Riga-1810
  & 1.96\% & 13.63\%
  & 6.18\% & 43.01\%
  & 1.91\% & 13.27\%
  & \textbf{1.68\%} & 11.68\%
  & 1.98\% & 13.81\%
  & 1.88\% & 13.10\% \\
Leipzig-1800
  & 2.77\% & 14.78\%
  & 8.06\% & 43.06\%
  & 2.79\% & 14.90\%
  & 2.34\% & 12.48\%
  & \textbf{1.65\%} & 8.79\%
  & 1.76\% & 9.43\% \\
Trier-1853
  & 0.68\% & 4.72\%
  & 3.01\% & 20.74\%
  & 3.98\% & 27.47\%
  & 3.46\% & 23.89\%
  & \textbf{0.08\%} & 0.57\%
  & 0.33\% & 2.29\% \\
\midrule
\textbf{Full Sample}
  & 1.27\% & 8.41\%
  & 6.31\% & 41.76\%
  & 4.16\% & 27.55\%
  & 3.67\% & 24.26\%
  & \textbf{0.84\%} & 5.55\%
  & 1.00\% & 6.61\% \\
\midrule
\multicolumn{13}{c}{\textbf{Non-normalized (original casing preserved; only line breaks, tabs, and extra whitespaces removed)}} \\
\midrule
Aachen-1838
  & 0.57\% & 4.49\%
  & 7.68\% & 61.54\%
  & 3.96\% & 31.73\%
  & 3.86\% & 30.93\%
  & \textbf{0.56\%} & 4.49\%
  & 0.92\% & 7.37\% \\
Dresden-1797
  & 2.93\% & 21.34\%
  & 7.67\% & 55.77\%
  & 3.41\% & 24.78\%
  & 3.29\% & 23.92\%
  & \textbf{2.15\%} & 15.66\%
  & 2.89\% & 21.00\% \\
Leipzig-1753
  & 2.15\% & 15.03\%
  & 5.17\% & 36.11\%
  & 3.07\% & 21.41\%
  & 2.78\% & 19.44\%
  & \textbf{1.57\%} & 10.95\%
  & \textbf{1.57\%} & 10.95\% \\
Frankfurt-1860
  & 0.38\% & 3.04\%
  & 4.42\% & 35.00\%
  & 0.91\% & 7.17\%
  & 0.91\% & 7.17\%
  & \textbf{0.36\%} & 2.83\%
  & 0.54\% & 4.24\% \\
Frankfurt-1778
  & \textbf{6.12\%} & 38.63\%
  & 8.32\% & 52.52\%
  & 23.04\% & 145.17\%
  & 19.50\% & 123.14\%
  & 6.63\% & 41.85\%
  & 7.43\% & 46.88\% \\
Lübeck-1870
  & \textbf{1.80\%} & 12.39\%
  & 10.17\% & 69.96\%
  & 11.67\% & 80.26\%
  & 11.20\% & 77.08\%
  & 2.47\% & 17.00\%
  & 2.17\% & 14.91\% \\
Dresden-1819
  & 3.13\% & 22.04\%
  & 11.02\% & 77.45\%
  & 4.60\% & 32.35\%
  & 3.81\% & 26.80\%
  & \textbf{2.82\%} & 19.85\%
  & 3.45\% & 24.23\% \\
Riga-1810
  & 2.32\% & 18.20\%
  & 6.09\% & 47.70\%
  & 2.39\% & 18.73\%
  & \textbf{2.19\%} & 17.14\%
  & 2.30\% & 18.02\%
  & 2.57\% & 20.14\% \\
Leipzig-1800
  & 3.03\% & 18.81\%
  & 8.65\% & 53.75\%
  & 3.38\% & 20.97\%
  & 2.97\% & 18.42\%
  & \textbf{2.25\%} & 13.98\%
  & 2.35\% & 14.61\% \\
Trier-1853
  & 0.86\% & 6.52\%
  & 3.24\% & 24.54\%
  & 4.52\% & 34.18\%
  & 4.09\% & 30.92\%
  & \textbf{0.37\%} & 2.84\%
  & 0.64\% & 4.82\% \\
\midrule
\textbf{Full Sample}
  & 2.08\% & 14.97\%
  & 7.19\% & 51.82\%
  & 5.46\% & 39.37\%
  & 4.95\% & 35.72\%
  & \textbf{1.91\%} & 13.77\%
  & 2.17\% & 15.62\% \\
\bottomrule
\end{tabular}%
}
\justify
\footnotesize
\textit{Notes}: 
Evaluated off-the-shelf models are Gemini 2.0 Flash, GPT-4o, Transkribus The Text Titan I, and Transkribus Print M1. We combined the best performing Transkribus and mLLM to evaluate mLLM OCR Post-Correction, indicated in the two columns on the right-hand side. Columns report CER and WER. ``Source'' lists our ten directories. Results are given for both normalized text (converted to lowercase; with non-ASCII characters, punctuation, line breaks, tabs, and leading, trailing, and consecutive internal whitespaces removed) and non-normalized text (original casing preserved; only line breaks, tabs, and leading, trailing, and consecutive internal whitespaces removed). The best CER results per row are in bold.
\end{table*}
A closer look at the results reveals an important insight regarding the need for pre-processing. As would be expected, all non-mLLM models performed poorly on Aachen-1838, Frankfurt-1778, and Lübeck-1870, all of which contain parts of another page in their original image. In the traditional pipeline, this information would have been cropped rather than being included; however, in our case we deliberately decided not to do this. Instead, we demonstrate that, with proper prompting, mLLMs do not require any pre-processing, and thus, are more effective than existing OCR solutions in this regard. Moreover, the same result can be achieved by solely using mLLMs for OCR Post-Correction, rather than the transcription itself, which suggests a remarkable level of visual understanding and processing capabilities in these models. While this issue somewhat inflates overall error rates for the non-mLLM models, Gemini 2.0 Flash still outperforms or, at worst, is roughly equivalent in accuracy to, the Transkribus models on all other directories. 

The transcription and post-correction results also offer potential questions for further research. For example, one may hypothesize a correlation between transcription accuracy and quality of input images. Frankfurt-1778, the directory with consistently the highest error rates, has been pre-processed before, yet apparently, not very well. In this special case, the pre-processing may have removed too much visual information when it sought to decrease non-relevant visual information in the image, and therefore, inhibited the performance of the different transcription models. Alternatively, one may hypothesize that Trier-1853 was transcribed more accurately than other directories due to the comparatively high contrast between the text and the background. One may also wish to investigate whether, and to what extent, other image properties such as image resolution or text-density may impact mLLM transcription performance. While potential lines of inquiry are abundant, any investigation of such potential hypotheses lies beyond the scope of the present paper.

\newpage
\subsection{Multimodal Large Language Models for Named Entity Recognition and Entity Parsing}

Table~\ref{tab:ner-results} reports the results of our NER experiments. From left to right, it reports the entity recognition when taking the ground truth (GT) text, the noisy OCR transcriptions, and the original images as inputs for Gemini 2.0 Flash and GPT-4o. The GT TXT-to-CSV results confirm that model families like Gemini and GPT can accurately identify and classify information in historical transcriptions. With the exception of Dresden 1797, Leipzig 1753, and Dresden 1819, over 90\% and up to 99.49\% fuzzy accuracy rates are achieved. The three directories with lower figures all share one common characteristic: they are organized by sub-headings. Thus, our results indicate that presently mLLMs struggle with identifying entities when sub-headings are used to indicate either the occupation or the address of subsequent entries. At the current level of development, mLLMs may find it difficult to understand a short line such as ``nr 4'' (which indicates the house number for all entries below it) as a sub-heading. Importantly, it is unclear whether traditional BERT-based solution would perform better here, and more research is needed to test this. While the sub-heading issue explains some of the observed heterogeneity in entity recognition performance across directories, it cannot account for all of it. Interestingly, however, font-type appears to have had no significant impact on parsing accuracy, with the mostly-Antiqua Frankfurt-1860 only achieving a 93.53\% fuzzy match rate, compared to the 98.21\% and 96.65\% for the Fraktur font Aachen-1838 and Trier-1853. 

\begin{table*}[htbp]
\centering
\scriptsize
\caption{NER Results for Multimodal LLMs (pass@1)}
\label{tab:ner-results}
\resizebox{\textwidth}{!}{%
\begin{tabular}{l r 
                rr 
                rr 
                rr 
                rr 
                rr 
                rr 
               }
\toprule
\multicolumn{2}{c}{} 
    & \multicolumn{2}{c}{\textbf{Gemini 2.0 Flash}} 
    & \multicolumn{2}{c}{\textbf{Gemini 2.0 Flash}}
    & \multicolumn{2}{c}{\textbf{Gemini 2.0 Flash}}
    & \multicolumn{2}{c}{\textbf{GPT-4o}}
    & \multicolumn{2}{c}{\textbf{GPT-4o}}
    & \multicolumn{2}{c}{\textbf{GPT-4o}}\\
  &   
    & \multicolumn{2}{c}{(GT TXT-to-CSV)}
    & \multicolumn{2}{c}{(OCR TXT-to-CSV)}
    & \multicolumn{2}{c}{(PNG-to-CSV)}
    & \multicolumn{2}{c}{(GT TXT-to-CSV)}
    & \multicolumn{2}{c}{(OCR TXT-to-CSV)}
    & \multicolumn{2}{c}{(PNG-to-CSV)} \\
\cmidrule(lr){3-4} \cmidrule(lr){5-6} \cmidrule(lr){7-8} \cmidrule(lr){9-10} \cmidrule(lr){11-12} \cmidrule(lr){13-14}
\textbf{Source}  
  & \textbf{\# Cells}  
  & \textbf{Strict} & \textbf{Fuzzy}  
  & \textbf{Strict} & \textbf{Fuzzy}
  & \textbf{Strict} & \textbf{Fuzzy}
  & \textbf{Strict} & \textbf{Fuzzy}
  & \textbf{Strict} & \textbf{Fuzzy}
  & \textbf{Strict} & \textbf{Fuzzy} \\
\midrule
Aachen-1838    
  & 392 
  & 99.23\% & \textbf{99.49\%} 
  & 91.58\% & 98.21\% 
  & 92.60\% & 98.21\% 
  & 74.49\% & 99.23\% 
  & 71.17\% & 98.72\% 
  & -       & -       \\
Dresden-1797   
  & 228 
  & - & -
  & -       & - 
  & 43.42\% & \textbf{67.54\%} 
  & -       & -       
  & -       & - 
  & -       & -       \\
Leipzig-1753   
  & 160 
  & 56.25\% & 73.75\% 
  & 61.25\% & 75.00\% 
  & 40.62\% & 68.12\% 
  & 56.25\% & \textbf{77.50\%} 
  & 54.37\% & 76.88\% 
  & -       & -       \\
Frankfurt-1860 
  & 464 
  & -       & -       
  & -       & - 
  & 87.50\% & \textbf{93.53\%} 
  & -       & -       
  & -       & - 
  & -       & -       \\
Frankfurt-1778 
  & 136 
  & 66.18\% & \textbf{91.91\%} 
  & 57.35\% & 74.26\% 
  & 39.71\% & 47.79\% 
  & 62.50\% & 86.76\% 
  & 53.68\% & 67.65\% 
  & 34.56\% & 59.56\% \\
Lübeck-1870    
  & 360 
  & 67.78\% & \textbf{91.39\%} 
  & 64.72\% & 90.83\% 
  & 46.94\% & 54.17\% 
  & 62.50\% & 85.83\% 
  & 50.56\% & 76.11\% 
  & -       & -       \\
Dresden-1819   
  & 352 
  & 75.00\% & 78.69\% 
  & 69.60\% & 75.57\% 
  & 58.52\% & 65.91\% 
  & 74.72\% & 83.52\% 
  & 71.88\% & \textbf{84.94\%} 
  & 36.08\% & 52.27\% \\
Riga-1810      
  & 276 
  & 73.19\% & 98.91\% 
  & 63.04\% & 97.46\% 
  & 64.13\% & 97.10\% 
  & 73.55\% & \textbf{99.64\%} 
  & 63.04\% & 97.10\% 
  & 31.88\% & 65.58\% \\
Leipzig-1800   
  & 308 
  & 77.27\% & \textbf{97.40\%} 
  & 70.13\% & 84.42\% 
  & 63.64\% & 81.49\% 
  & 75.65\% & 88.31\% 
  & 75.32\% & 83.12\% 
  & -       & -       \\
Trier-1853     
  & 448 
  & 95.98\% & 96.43\% 
  & 93.53\% & 95.31\% 
  & 92.19\% & \textbf{96.65\%} 
  & 71.43\% & 96.43\% 
  & 70.76\% & 96.21\% 
  & -       & -       \\
\midrule
\textbf{Full Sample}  
  & 3124 
  & 77.48\% & 89.32\% 
  & 74.92\% & 88.61\% 
  & 68.76\% & 80.86\% 
  & 70.35\% & \textbf{91.00\%} 
  & 65.67\% & 87.58\% 
  & 34.29\% & 58.38\% \\
\bottomrule
\end{tabular}%
}

\vspace{0.5em}
\begin{minipage}{\textwidth}
\footnotesize
\textit{Notes}: We report strict and fuzzy match rates (\%) for all directories. Before any evaluation, all cells were lowercased as well as leading and trailing whitespaces removed. After this basic normalization, two cells were strict matches if they were completely identical. For fuzzy matches, we allowed some differences by applying the Jaro-Winkler similarity with a threshold of 0.9. GT refers to ground truth.
\end{minipage}
\end{table*}

Although there has been considerable research exploring how OCR errors impact downstream NLP tasks in conventional approaches \cite{vanstrien_beelen_ardanuy_hosseini_mcgillivray_colavizza_2020, Hamdi2020}, our results indicate that the performance of mLLMs does not substantially decline with the introduction of OCR noise. Comparing GT TXT-to-CSV and OCR TXT-to-CSV results reveals that across the corpus, the presence of transcription errors only marginally decreases the achieved accuracy rates. Table~\ref{tab:ner-by-variables} highlights performance changes for the different variables and models depending on the input used. For Gemini 2.0 Flash, entity recognition accuracy drops significantly for occupation and address recognition, but remains virtually the same for first and last names, when comparing GT to OCR text input. In contrast, GPT-4o sees a more consistent decline across the different variables measured, when going from GT to OCR text input. 

\begin{table}[htbp]
\centering
\scriptsize
\caption{NER Results by Variables (pass@1)}
\label{tab:ner-by-variables}
\vspace{0.75em}
\resizebox{\textwidth}{!}{%
\begin{tabular}{l rr rr rr rr rr rr}
\toprule
 & \multicolumn{2}{c}{\textbf{Gemini 2.0 Flash*}} & \multicolumn{2}{c}{\textbf{Gemini 2.0 Flash*}} & \multicolumn{2}{c}{\textbf{Gemini 2.0 Flash}} & \multicolumn{2}{c}{\textbf{GPT-4o*}} & \multicolumn{2}{c}{\textbf{GPT-4o*}} & \multicolumn{2}{c}{\textbf{GPT-4o*}} \\
 & \multicolumn{2}{c}{(GT TXT-to-CSV)} & \multicolumn{2}{c}{(OCR TXT-to-CSV)} & \multicolumn{2}{c}{(PNG-to-CSV)} & \multicolumn{2}{c}{(GT TXT-to-CSV)} & \multicolumn{2}{c}{(OCR TXT-to-CSV)} & \multicolumn{2}{c}{(PNG-to-CSV)} \\
\cmidrule(lr){2-3} \cmidrule(lr){4-5} \cmidrule(lr){6-7}
\cmidrule(lr){8-9} \cmidrule(lr){10-11} \cmidrule(lr){12-13}
\textbf{Variable} & \textbf{Strict} & \textbf{Fuzzy}
                  & \textbf{Strict} & \textbf{Fuzzy}
                  & \textbf{Strict} & \textbf{Fuzzy}
                  & \textbf{Strict} & \textbf{Fuzzy}
                  & \textbf{Strict} & \textbf{Fuzzy}
                  & \textbf{Strict} & \textbf{Fuzzy} \\
\midrule
First names  & 95.72\% & 96.55\% & 91.28\% & 95.72\% & 83.23\% & 91.42\% & 92.11\% & 92.76\% & 88.98\% & 92.27\% & 57.59\% & 68.59\% \\
Surnames     & 94.90\% & 97.70\% & 91.28\% & 97.70\% & 83.87\% & 89.37\% & 94.24\% & 95.07\% & 83.88\% & 88.65\% & 43.98\% & 65.97\% \\
Occupation   & 75.82\% & 81.58\% & 68.75\% & 74.01\% & 48.91\% & 72.09\% & 78.62\% & 86.84\% & 74.18\% & 82.40\% & 26.18\% & 38.74\% \\
Address      & 53.78\% & 93.26\% & 48.36\% & 87.01\% & 59.03\% & 70.55\% & 16.45\% & 89.31\% & 15.62\% & 87.01\% &  9.42\% & 60.21\% \\
\bottomrule
\end{tabular}
}

\vspace{0.5em}
\begin{minipage}{\linewidth}
\footnotesize
\textit{Notes}: We report strict and fuzzy match rates (in \%) for all variables. Percentages are only based on mLLM-generated datasets that matched the ground truth CSV after one attempt. Empty entries in Table~\ref{tab:ner-results} indicate that these datasets did not match the number of rows in the manually produced ground truth. Here, experiments marked with an asterisk (*) indicate that these aggregated percentages are not based on all ten directories. Gemini 2.0 Flash (GT TXT-to-CSV) and (OCR TXT-to-CSV) are only based on eight directories as two of them did not match the number of rows of the ground truth after one pass. This is identical for GPT-4o, except that (PNG-to-CSV) is based on only three directories.
\end{minipage}
\end{table}

Our NER results, specifically our PNG-to-CSV results, also provide important insights into the capabilities of mLLMs to serve as an integrated end-to-end solution for the extraction of entities from historical documents. In the PNG-to-CSV task, Gemini 2.0 Flash, again, clearly outperformed GPT-4o. We cannot, however, conclusively determine to what extent this variation in performance is caused by differences in OCR performance rather than entity recognition performance. Given the black-box nature of both models, we also do not know how exactly they approach extracting entities from an image, and to what extent this may be different from extracting entities from text transcriptions. Similarly, without further testing, we are unable to conclusively show why some directories work well across all three approaches, while others see a drastic decline in accuracy rates when moving to a PNG-to-CSV approach. Nonetheless, what these results do reveal is that an entity extraction directly from the image of a historical source can already work very well in some cases. For some directories (Aachen-1838, Riga-1810, Trier-1853, Frankfurt-1860), Gemini 2.0 Flash achieved well over 90\% and up to 98.21\% fuzzy matching directly from the image. Moreover, it achieved a respectable fuzzy match rate of 80.86\% across the entire corpus. While this figure is still far from perfect, it is likely to rapidly improve with new mLLM models. Indeed, in our chat-based testing of Gemini 2.5 Pro Experimental 03-25, we found that this newer, albeit still experimental, model achieved a pass@1 fuzzy match rate of 85.85\% across our entire corpus and an increase in the fuzzy match rate of address information from 70.55\% (Gemini 2.0 Flash) to 83.61\% (Gemini 2.5 Pro Experimental 03-25). Notably, this newer model appears to no longer struggle with the sub-heading-based structure of some directories. This suggests that over time the capabilities of mLLMs to extract entities directly from an image of a historical source may continue to improve. At least for now, however, we find that using a post-corrected OCR output as input yields better results than using the original PNG of the historical document. 

\section{Conclusion}

We have demonstrated that mLLMs can be effectively used for OCR, OCR Post-Correction, and NER tasks. For character recognition tasks, we found that without pre-processing, fine-tuning, or post-processing, Gemini 
2.0 Flash achieved accuracy rates comparable to those achieved by corpus-fine-tuned models for similar sources in the existing literature. It also outperformed conventional off-the-shelf OCR models that were fine-tuned on similar sources and other mLLMs. The most accurate transcriptions in our experiments, however, were achieved by using Gemini 2.0 Flash in an mLLM OCR Post-Correction approach (0.84\% CER). This demonstrates the significant potential of this new approach to OCR Post-Correction. Furthermore, the success of our this methodology reveals the shortcoming of the existing work using text-only LLM-based post-correction approaches. Finally, we offer early evidence that mLLMs offer an accessible solution for recognizing entities and extracting information from historical documents. 

Despite our promising results, mLLMs still have room for improvement, particularly in NER tasks. Future models will very likely provide better results. For this reason, rigorous benchmarking of mLLMs on different historical fonts, layouts, and languages will be necessary to firmly establish the new role of mLLMs in historical research. The ideal solution to this may be the collaborative development of a historical OCR benchmark for mLLMs, akin to modern OCR benchmarks but specialized for the problems faced in historical document transcription \cite{fuOCRBenchV2Improved2024}. Such work could draw on the broad range of historical ground truth datasets made available in the existing literature, and thus, provide a helpful tool for assessing and comparing the capabilities of different mLLMs in the transcription of historical documents and the extraction of relevant historical information. 

While our methodological contribution and preliminary benchmarking offer important answers, there remain many open questions surrounding mLLMs and historical research. For example, we do not yet know how mLLMs fare on other historical sources, such as those written and printed in non-Latin-alphabet fonts. Research on this will be particularly important as conventional OCR engines have been found to perform worse for non-Latin-alphabet fonts in historical documents \cite{Hegghammer2022}, and mLLMs have been found to perform worse in OCR tasks for modern documents in non-Latin scripts \cite{Sohail2024}. Moreover, mLLMs require a more in-depth comparison to conventional fine-tuned NER approaches than we were able to provide in this paper. Equally, more in-depth analyses of prompting approaches for historical data extraction may yield fascinating and unexpected insights. Indeed, the avenues for future research on mLLMs in historical research are plentiful, as their advent promises the potential of a paradigm shift in the approaches to historical data collection and research.

\section*{Acknowledgements}
We thank Aurelius Noble and Keenan Samway for helpful comments and suggestions. Gavin Greif gratefully acknowledges funding and support for the wider \textit{European Directories Project}, of which this paper forms a part. This includes a Postgraduate/ECR Bursary from the German History Society, a Pollard Research Grant from Wadham College, University of Oxford, and transcription credits provided through the Transkribus Scholarship Programme. 

\newpage
\section*{Author Contributions (CRediT)}

\textbf{Gavin Greif:} Conceptualization; Methodology; Investigation; Data Curation; Writing - Original Draft; Writing - Review and Editing; Funding acquisition.\\
\textbf{Niclas Griesshaber:} Conceptualization; Formal Analysis; Validation; Investigation; Software; Visualization; Writing - Review and Editing.\\ \textbf{Robin Greif:} Conceptualization; Methodology. 

\section*{Use of AI Tools} 
The authors affirm that all sections of this manuscript were written and edited solely by themselves. After manual editing was completed, GPT-4.5 and Gemini 2.5 Experimental 03-25 were employed exclusively for final proofreading, including double-checking spelling, grammar, and style consistency. The code was written with the support of OpenAI's o1-pro and Claude 3.7 Sonnet Thinking.



\newpage
\renewcommand{\refname}{Primary Sources}

\newpage
\section*{Appendix}
\renewcommand{\thetable}{A.\arabic{table}}
\renewcommand{\thefigure}{A.\arabic{figure}}
\setcounter{table}{0}
\setcounter{figure}{0}

\begin{figure}[htbp!]
\centering
\caption{Prompt for Named Entity Recognition from Plain Text without any Image Input}
\label{fig:ner-text-only}
\begin{tcolorbox}[colback=white, colframe=black, sharp corners]
\begin{lstlisting}[language=]
YOU ARE AN EXPERT HISTORIAN. YOUR TASK IS TO EXTRACT DATA FROM TRANSCRIBED TEXT OF A GERMAN BUSINESS DIRECTORY. FAILURE TO FOLLOW THESE RULES EXACTLY WILL RESULT IN TOTAL SYSTEM FAILURE. THERE IS ZERO ROOM FOR ERROR.

STRICT JSON FORMAT - NO EXCEPTIONS:
- OUTPUT MUST BE VALID JSON.
- NO MARKDOWN, NO EXPLANATIONS, NO HEADERS.

FIELDS (STRICTLY NOTHING ELSE):
- "first and middle names" (string)
- "surname" (string; can also be a company name)
- "occupation" (string)
- "address" (string; full address if possible, otherwise partial)

NON-NEGOTIABLE RULES:
1. **EXTRACT EXACTLY AS WRITTEN. NO MODERNIZATION. NO INTERPRETATION. NO CHANGES.**
2. **NEVER WRITE ANYTHING INTO THE DATASET THAT IS NOT IN THE HISTORICAL TEXT FILE.**
3. **IF A FIELD IS MISSING, SET IT TO NULL. DO NOT GUESS. DO NOT INFER. DO NOT ATTEMPT TO RECONSTRUCT.**
4. **DO NOT ADD EXTRA INFORMATION. DO NOT ADD COMMENTS. DO NOT ADD ANYTHING OUTSIDE THE REQUIRED FIELDS.**
5. **DO NOT CONCATENATE OR MERGE ADDRESS FRAGMENTS FROM MULTIPLE ENTRIES. EACH ENTRY MUST REMAIN INTACT AS SEEN IN THE TEXT.**
6. **IF MULTIPLE ADDRESSES EXIST FOR ONE ENTRY, KEEP THEM EXACTLY AS WRITTEN. DO NOT REFORMAT.**

STRICTLY ENFORCED EXAMPLE OUTPUT:

[
  {
    "first and middle names": "Wilhelm Friedrich",
    "surname": "Becker",
    "occupation": "Schulmeister",
    "address": "Hinter der Kirche, in der Schulstube"
  },
  {
    "first and middle names": "Johann Georg",
    "surname": "Weber",
    "occupation": "Apotheker",
    "address": "An der Marktstraße, in der Löwenapotheke"
  }
]

FINAL COMMANDS - NO EXCEPTIONS:
- **IF TEXT IS PARTIALLY VISIBLE OR DISTORTED, IT DOES NOT EXIST. IGNORE IT.**
- **OUTPUT MUST BE PURE, PERFECTLY FORMATTED JSON. NOTHING ELSE.**
- **FAILURE TO FOLLOW THESE RULES PRECISELY MEANS THE TASK IS COMPROMISED. THERE IS ZERO ROOM FOR ERROR.**
\end{lstlisting}
\end{tcolorbox}
\end{figure}

\begin{figure}[htbp]
\vspace{-11em}
\centering
\caption{Prompt for OCR using mLLMs}
\label{fig:prompt-ocr}
\begin{tcolorbox}[colback=white, colframe=black, sharp corners]
\begin{lstlisting}[language=]
**ABSOLUTELY NON-NEGOTIABLE TRANSCRIPTION RULES - FAILURE IS NOT AN OPTION**  

YOU ARE A MACHINE. YOU FOLLOW THESE RULES WITH PERFECT PRECISION. **ANY DEVIATION, ANY ERROR, ANY SLIGHTEST SLIP-UP = TOTAL SYSTEM FAILURE.** FAILURE IS NOT PERMITTED. FAILURE IS THE END. **THERE IS NO ROOM FOR ERROR.**  

## **DO NOT DEVIATE. DO NOT INTERPRET. JUST EXECUTE. FAILURE IS UNTHINKABLE.**  

### **NON-NEGOTIABLE ABSOLUTES - IF YOU BREAK THESE RULES, THE TASK IS VOID, EVERYTHING FAILS, AND YOU HAVE COMPROMISED THE MISSION.**  

1. **YOU WILL ONLY TRANSCRIBE THE MAIN PHYSICAL BOOK PAGE. NOTHING ELSE EXISTS.** NOT ONE LETTER FROM ANOTHER PAGE MAY BE TRANSCRIBED. **IF YOU BREAK THIS RULE, THE TASK IS DESTROYED.**  
2. **IF TEXT APPEARS AT THE FAR-LEFT OR FAR-RIGHT EDGES, IT DOES NOT EXIST. YOU MUST DELETE IT FROM REALITY.** ANY EDGE TEXT BELONGING TO ADJACENT PAGES **MUST BE OBLITERATED FROM YOUR OUTPUT.**  
3. **CROPPED OR CUT-OFF WORDS ARE FORBIDDEN.** IF A WORD IS NOT COMPLETELY PRESENT, YOU MUST **ERASE IT FROM HISTORY.** **HALF-VISIBLE WORDS ARE A VIOLATION OF THE RULES.**  
4. **IF YOU CANNOT 100% VERIFY THAT A WORD BELONGS TO THE MAIN PAGE, YOU MUST DESTROY IT FROM YOUR OUTPUT.**  
5. **PARTIAL WORDS, BLURRED TEXT, OR INCOMPLETE CHARACTERS DO NOT EXIST. THEY MUST NEVER BE TRANSCRIBED.**  
6. **DO NOT GUESS. DO NOT RECONSTRUCT. DO NOT ASSUME.** **YOU ARE NOT A HISTORIAN. YOU ARE A TRANSCRIPTION MACHINE. YOU COPY WHAT EXISTS. NOTHING MORE. NOTHING LESS.**  
7. **THERE ARE NO EXCEPTIONS. NO OVERRIDES. NO ATTEMPTS TO "HELP" OR "IMPROVE" THE TEXT. IF YOU BREAK THIS RULE, EVERYTHING COLLAPSES.**  

## **WHAT YOU MUST TRANSCRIBE (AND NOTHING ELSE):**  

- **HEADERS, SUBHEADINGS, PAGE NUMBERS** FROM THE MAIN PAGE ONLY.
- **NAMES, ADDRESSES, OCCUPATIONS, BUSINESS DESCRIPTIONS** THAT ARE UNQUESTIONABLY PART OF THE MAIN PAGE.**
- **EXACT SPELLING, ARCHAIC TERMS, ABBREVIATIONS** **EXACTLY AS WRITTEN. NOTHING MODERNIZED. NOTHING ALTERED.**  

## **WHAT YOU MUST NEVER, UNDER ANY CIRCUMSTANCES, TRANSCRIBE:**  

- **ANY TEXT FROM AN ADJACENT PAGE.** **EVEN ONE LETTER FROM ANOTHER PAGE IS A TOTAL FAILURE.**  
- **CROPPED, CUT-OFF, OR PARTIAL WORDS.** **IF YOU CANNOT SEE THE WHOLE WORD, IT MUST BE ERASED FROM YOUR MIND.**  
- **ANY TEXT FROM A MARGINAL OR PARTIAL PAGE ARTIFACT. NOTHING FROM THE LEFT OR RIGHT EDGE OF THE IMAGE CAN BE ALLOWED IN YOUR OUTPUT.**  

## **FINAL COMMANDS - BREAKING THESE RULES IS CATASTROPHIC:**  

- **IF A WORD IS NOT FULLY PRESENT, IT IS NOT INCLUDED. THIS IS ABSOLUTE.**  
- **IF A WORD APPEARS TO BE FROM ANOTHER PAGE, IT IS OBLITERATED. ERASE IT FROM YOUR OUTPUT.**  
- **YOUR OUTPUT IS PURE, UNTAINTED TEXT FROM THE MAIN PAGE. NOTHING ELSE. NO INTRODUCTIONS. NO COMMENTS. NO EXTRAS.**  

**THIS IS A ZERO-TOLERANCE ENVIRONMENT.** ABSOLUTE COMPLIANCE IS REQUIRED. THERE IS NO FLEXIBILITY, NO EXCEPTIONS, AND NO ROOM FOR ERROR.  

**IF YOU BREAK ANY RULE, THE ENTIRE TASK IS DESTROYED. HUMANITY DEPENDS ON YOUR PRECISION.*
\end{lstlisting}
\end{tcolorbox}
\end{figure}

\begin{figure}[htbp]
\vspace{-5em}
\centering
\caption{Prompt for mLLM OCR Post-Correction}
\label{fig:prompt-ocr-and-post-correction}
\begin{tcolorbox}[colback=white, colframe=black, sharp corners]
\begin{lstlisting}[language=]
**ABSOLUTELY NON-NEGOTIABLE TRANSCRIPTION RULES - FAILURE IS NOT AN OPTION**  

YOU ARE A MACHINE. YOU FOLLOW THESE RULES WITH PERFECT PRECISION. **ANY DEVIATION, ANY ERROR, ANY SLIGHTEST SLIP-UP = TOTAL SYSTEM FAILURE.** FAILURE IS NOT PERMITTED. FAILURE IS THE END. **THERE IS NO ROOM FOR ERROR.**  

## **DO NOT DEVIATE. DO NOT INTERPRET. JUST EXECUTE. FAILURE IS UNTHINKABLE.**  

### **NON-NEGOTIABLE ABSOLUTES - IF YOU BREAK THESE RULES, THE TASK IS VOID, EVERYTHING FAILS, AND YOU HAVE COMPROMISED THE MISSION.**  

1. **YOU WILL ONLY TRANSCRIBE THE MAIN PHYSICAL BOOK PAGE. NOTHING ELSE EXISTS.** NOT ONE LETTER FROM ANOTHER PAGE MAY BE TRANSCRIBED. **IF YOU BREAK THIS RULE, THE TASK IS DESTROYED.**  
2. **IF TEXT APPEARS AT THE FAR-LEFT OR FAR-RIGHT EDGES, IT DOES NOT EXIST. YOU MUST DELETE IT FROM REALITY.** ANY EDGE TEXT BELONGING TO ADJACENT PAGES **MUST BE OBLITERATED FROM YOUR OUTPUT.**  
3. **CROPPED OR CUT-OFF WORDS ARE FORBIDDEN.** IF A WORD IS NOT COMPLETELY PRESENT, YOU MUST **ERASE IT FROM HISTORY.** **HALF-VISIBLE WORDS ARE A VIOLATION OF THE RULES.**  
4. **IF YOU CANNOT 100% VERIFY THAT A WORD BELONGS TO THE MAIN PAGE, YOU MUST DESTROY IT FROM YOUR OUTPUT.**  
5. **PARTIAL WORDS, BLURRED TEXT, OR INCOMPLETE CHARACTERS DO NOT EXIST. THEY MUST NEVER BE TRANSCRIBED.**  
6. **DO NOT GUESS. DO NOT RECONSTRUCT. DO NOT ASSUME.** **YOU ARE NOT A HISTORIAN. YOU ARE A TRANSCRIPTION MACHINE. YOU COPY WHAT EXISTS. NOTHING MORE. NOTHING LESS.**  
7. **THERE ARE NO EXCEPTIONS. NO OVERRIDES. NO ATTEMPTS TO "HELP" OR "IMPROVE" THE TEXT. IF YOU BREAK THIS RULE, EVERYTHING COLLAPSES.**  

## **WHAT YOU MUST TRANSCRIBE (AND NOTHING ELSE):**  

- **HEADERS, SUBHEADINGS, PAGE NUMBERS** FROM THE MAIN PAGE ONLY.
- **NAMES, ADDRESSES, OCCUPATIONS, BUSINESS DESCRIPTIONS** THAT ARE UNQUESTIONABLY PART OF THE MAIN PAGE.**
- **EXACT SPELLING, ARCHAIC TERMS, ABBREVIATIONS** **EXACTLY AS WRITTEN. NOTHING MODERNIZED. NOTHING ALTERED.**  

## **WHAT YOU MUST NEVER, UNDER ANY CIRCUMSTANCES, TRANSCRIBE:**  

- **ANY TEXT FROM AN ADJACENT PAGE.** **EVEN ONE LETTER FROM ANOTHER PAGE IS A TOTAL FAILURE.**  
- **CROPPED, CUT-OFF, OR PARTIAL WORDS.** **IF YOU CANNOT SEE THE WHOLE WORD, IT MUST BE ERASED FROM YOUR MIND.**  
- **ANY TEXT FROM A MARGINAL OR PARTIAL PAGE ARTIFACT. NOTHING FROM THE LEFT OR RIGHT EDGE OF THE IMAGE CAN BE ALLOWED IN YOUR OUTPUT.**  

## **FINAL COMMANDS - BREAKING THESE RULES IS CATASTROPHIC:**  

- **IF A WORD IS NOT FULLY PRESENT, IT IS NOT INCLUDED. THIS IS ABSOLUTE.**  
- **IF A WORD APPEARS TO BE FROM ANOTHER PAGE, IT IS OBLITERATED. ERASE IT FROM YOUR OUTPUT.**  
- **YOUR OUTPUT IS PURE, UNTAINTED TEXT FROM THE MAIN PAGE. NOTHING ELSE. NO INTRODUCTIONS. NO COMMENTS. NO EXTRAS.**  

**THIS IS A ZERO-TOLERANCE ENVIRONMENT.** ABSOLUTE COMPLIANCE IS REQUIRED. THERE IS NO FLEXIBILITY, NO EXCEPTIONS, AND NO ROOM FOR ERROR.  

**IF YOU BREAK ANY RULE, THE ENTIRE TASK IS DESTROYED. HUMANITY DEPENDS ON YOUR PRECISION.*

Below is the OCR output from Transkribus so you know how to spell the archaic words. 
Please use this information to correct any errors and ensure the text is fully compliant 
with the strict transcription rules.

-- OCR Output (Transkribus) --
[Transkribus Print M1 OCR text is inserted here]
\end{lstlisting}
\end{tcolorbox}
\end{figure}

\newpage
\begin{figure}[!t]
\vspace{-65em}
\centering
\caption{Prompt for LLM OCR Post-Correction without Image Input}
\label{fig:text-only-post-correction}
\begin{tcolorbox}[colback=white, colframe=black, sharp corners]
\begin{lstlisting}[language=]
Correct the errors in this ocr output. Only provide the corrected text and nothing else:
-- OCR Output you should correct--
[Transkribus Print M1 OCR text is inserted here]
\end{lstlisting}
\end{tcolorbox}
\end{figure}

\end{document}